\documentclass[10pt,twocolumn,letterpaper]{article}

\usepackage{iccv}              
\def\ours{\textsc{Fix-CLIP}\xspace}

\usepackage[accsupp]{axessibility}
\usepackage{tabularray}
\usepackage{xcolor}
\usepackage{changes}
\usepackage{graphicx}
\usepackage{amsmath}
\usepackage{amssymb}
\usepackage{bbding}
\usepackage{utfsym}
\usepackage{booktabs}
\usepackage{enumitem}
\usepackage{microtype}
\usepackage[dvipsnames]{xcolor}
\usepackage{bm, bbm}
\usepackage{algorithmic}
\usepackage[linesnumbered,boxed,ruled,commentsnumbered]{algorithm2e}
\usepackage{multirow}

\usepackage{makecell}

\usepackage[accsupp]{axessibility}

\definecolor{iccvblue}{rgb}{0.21,0.49,0.74}
\usepackage[pagebackref,breaklinks,colorlinks,allcolors=iccvblue]{hyperref}

\def\paperID{2498} 
\def\confName{ICCV}
\def\confYear{2025}

\title{\ours: Dual-Branch Hierarchical Contrastive Learning via Synthetic Captions for Better Understanding of Long Text}

\author{Bingchao Wang$^{1*}$, Zhiwei Ning$^{1*}$, Jianyu Ding$^{1*}$, Xuanang Gao$^{1*}$, \\
Yin Li$^{2}$, Dongsheng Jiang$^{2}$, Jie Yang$^{1\dagger}$, Wei Liu$^{1 \dagger}$\\
\small{
$^1$Shanghai Jiao Tong University \tt\small \{bc\_wang, zwning, jianyuding, fangkuar, jieyang, weiliucv\}@sjtu.edu.cn}\\
\small{$^2$Huawei Inc. \tt\small \{liyin9, jiangdongsheng1\}@huawei.com}\\
}

\begin{document}
\maketitle
\begin{abstract}
CLIP has shown promising performance across many short-text tasks in a zero-shot manner. However, limited by the input length of the text encoder, CLIP struggles on under-stream tasks with long-text inputs ($>77$ tokens). To remedy this issue, we propose \ours, which includes three novel modules: (1) A dual-branch training pipeline that aligns short and long texts with masked and raw images, respectively, which boosts the long-text representation while preserving the short-text ability. (2) Multiple learnable regional prompts with unidirectional masks in Transformer layers for regional information extraction. (3) A hierarchical feature alignment module in the intermediate encoder layers to promote the consistency of multi-scale features. Furthermore, we collect 30M images and utilize existing MLLMs to synthesize long-text captions for training. Extensive experiments show that \ours achieves state-of-the-art performance on both long-text and short-text retrieval benchmarks. For downstream applications, we reveal that \ours's text encoder delivers promising performance in a plug-and-play manner for diffusion models with long-text input. The code is available at \url{https://github.com/bcwang-sjtu/Fix-CLIP}.

{
  \renewcommand{\thefootnote}%
    {\fnsymbol{footnote}}
  \footnotetext[0]{*Equal contribution, $\dagger$Corresponding authors. This work is partially supported by NSFC (No. 62376153, 62402318, 24Z990200676).} 
  }
\end{abstract} 

\vspace{-1em}
\section{Introduction}
\label{sec:intro}
CLIP~\cite{radford2021learning} has garnered significant performance across various open-vocabulary tasks. It is widely used as the backbone in Multi-modality Large Language Model (MLLM)~\cite{li2022blip, li2023blip, cai2024internlm2, wang2023image} and generative models~\cite{Rombach_Blattmann_Lorenz_Esser_Ommer_2022, ni2025wonderturbo, chen2025prismlayers, fan2025go}.

The success of CLIP is based on the large-scale web-based image-text pairs, which have extremely short effective text length~\cite{zhang2024long}. In fact, images often require dozens of sentences to describe their content adequately. However, CLIP can not understand long text inputs, which severely limits its application to MLLMs and text-to-image generation models. Recently, PixArt-$\alpha$~\cite{chen2023pixart} uses Flan-T5 as the text encoder to increase the length of input tokens from 77 to 120 and injects the obtained text features into DiT~\cite{peebles2023scalable} to alleviate the deficiency in long-text understanding. Following Long-CLIP~\cite{zhang2024long}, our model achieves stronger performance with an input length of 248.

\begin{figure}[t] 
	\centering  
	\includegraphics[width=0.85\linewidth]{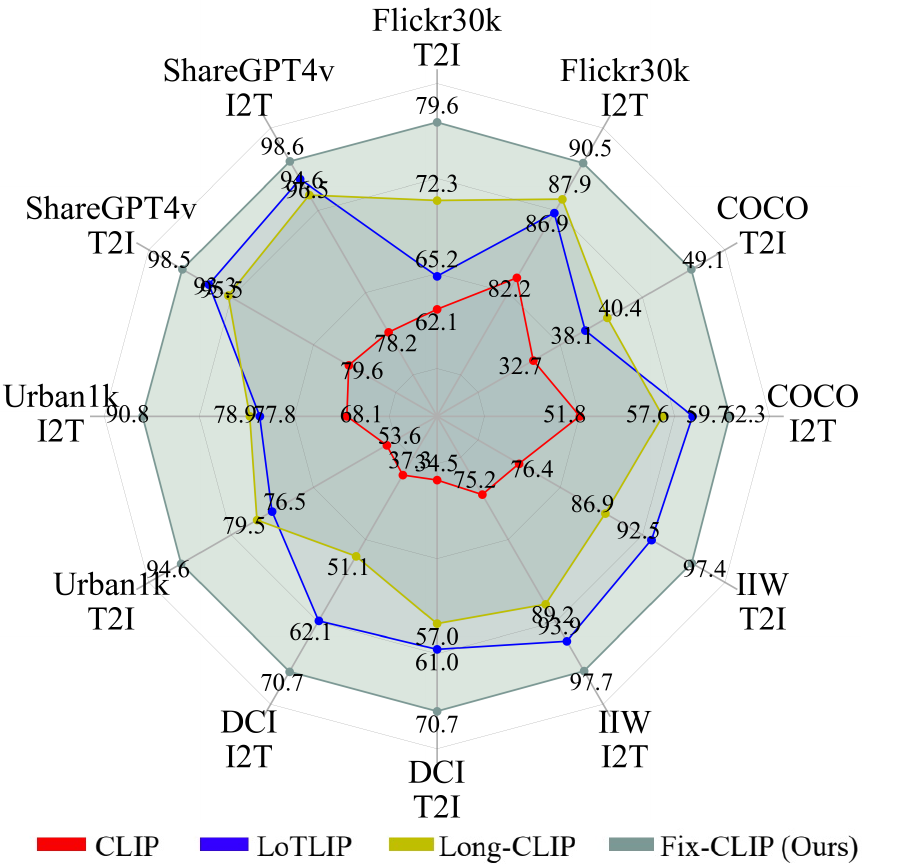}
        \vspace{-0.5em}
	\caption{We compare \ours with CLIP \cite{radford2021learning}, LoTLIP \cite{wu2024lotlip}, and Long-CLIP \cite{zhang2024long} on B/16 model. \ours achieves competitive performance across long-text and short-text retrieval tasks.}
        \vspace{-1em}
	\label{fig:retrieval}
\end{figure}

\begin{figure*}[t] 
	\centering  
	\includegraphics[width=1.0\linewidth]{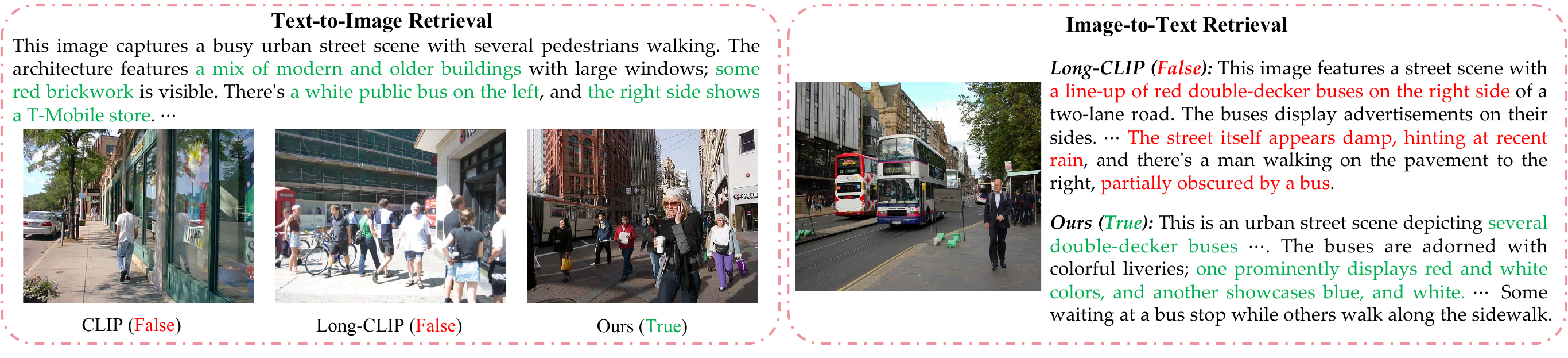}
        \vspace{-2em}
	\caption{Comparison of \ours against Long-CLIP \cite{zhang2024long} in image-to-text and text-to-image retrieval tasks with long-text captions. The key texts related to the correct elements are marked in green, and the red texts indicate the wrong elements.}
        \vspace{-1em}
	\label{fig:retrieval_examples}
\end{figure*}

To improve the understanding of long text, previous methods~\cite{zhang2024long, zheng2025dreamlip, wu2024lotlip} incorporate long caption datasets to enhance the alignment between image and long-text while maintaining the short-text performance through pre-training~\cite{zheng2025dreamlip, wu2024lotlip} or incremental training~\cite{zhang2024long} strategy. Nonetheless, the conventional training paradigm of contrastive learning aims to convert the [CLS] tokens of images and texts into a consistent feature space, which emphasizes global alignment rather than local alignment. The lack of local representation and long-text understanding leads to suboptimal performance on tasks requiring fine-grained description.

Due to the reason that effective extraction of image detail features is crucial, recent works~\cite{shi2025umg, wang2023position, abdollah2024comalign} make an effort to address the issue by dividing the input images into several regions and matching each region with the corresponding caption. These methods facilitate the detailed representation of image features explicitly. Furthermore, some methods~\cite{li2024cascade, mukhoti2023open, lan2024proxyclip} match patch embeddings in the image encoder midden layers with text features to implicitly enhance regional consistency. The explicit approaches need to generate corresponding captions for numerous image regions, leading to large data scales and high resource occupation. Conversely, methods that focus on implicit local consistency~\cite{li2024cascade,lan2024proxyclip} would inadvertently impact the generalization capability of the pre-trained model, resulting in the degraded performance of short-text tasks.


In this work, we optimize the implicit alignment strategy and conduct incremental training on the pre-trained model to achieve a balance between performance and resource consumption. We propose \ours to improve the understanding of long text and maintain the superior generalization ability in short-text tasks, as shown in \cref{fig:retrieval}. \cref{fig:retrieval_examples} visualizes our superiority over vanilla CLIP~\cite{radford2021learning} and Long-CLIP~\cite{zhang2024long} in image-text retrieval tasks.
The contributions of this paper are as follows:
\begin{itemize}
    \item A dual-branch training pipeline is proposed to align short and long texts with masked and raw images respectively. It enhances long-text capabilities while preventing the forgetting of CLIP's original short-text abilities.
    \item Regional prompts are designed  for better alignment between sub-texts and local visual features, assisted with a unidirectional mask to preserve the integrity of the patch embedding.
    \item A hierarchical feature alignment module is employed to promote the consistency of multi-scale features in the intermediate encoder layers, which optimizes contrastive learning in long texts.
    \item We instruct MLLMs to synthesize 30M long-text-image pairs for training. Our \ours achieves state-of-the-art performance on long-text and short-text benchmarks. The text encoder delivers promising performance in a plug-and-play manner for diffusion models with long-text input.
\end{itemize}

\section{Related Work}
\label{sec:related}


\subsection{Vision-Language Pre-training Model}
CLIP serial approaches~\cite{radford2021learning, cherti2023reproducible, jia2021scaling, xu2023demystifying, zhang2024vision} effectively mitigate the inconsistency in feature spaces between the output of the text encoder and the image encoder by restricted alignment strategy. As the pioneering works, CLIP~\cite{radford2021learning} and ALIGN~\cite{jia2021scaling} demonstrate that leveraging internet-sourced dataset (400M) enables promising results across computer vision tasks, including classification~\cite{sun2023eva, shi2025umg}, segmentation~\cite{zhang2024exploring, dong2023maskclip, zhou2023zegclip, li2022language, jain2024vcoder} and detection~\cite{simeoni2021localizing, gu2021open, li2022grounded}. The similar image and text encoder are designed to extract multi-modal information and project them into a shared space to achieve feature alignment. 

Benefiting from the generalization ability of CLIP, many subsequent methods~\cite{dong2023maskclip, DynamicID, you2025fbdifffourierbasisguideddiffusion, you2025temporaldifferentialfields4d, peng2025bizgen} achieve promising performance in open-world scenes. MaskCLIP~\cite{dong2023maskclip} and FLIP~\cite{li2023scaling} enhance the encoding capability by masking a large proportion of image patches. FILIP~\cite{yao2021filip} explores regional expressiveness by facilitating the consistency between patch tokens and text tokens. EVA-CLIP~\cite{sun2023eva} conducts novel techniques for stable and efficient training. However, the capability of long-text understanding remains the limitation of CLIP~\cite{radford2021learning}, which restricts the development of more complex vision-language applications.

\subsection{Long Text Understanding}
For computational efficiency, the sequence length is capped at 77 in CLIP, which prevents the subsequent information in long text. An image usually contains rich information and requires a lengthy caption to be described. In recent works, instructing LLMs or MLLMs to synthesize data has become a cost-effective choice for data synthesis. LaCLIP~\cite{fan2024improving} directly rewrites the original text descriptions through LLMs, which leads to serious hallucinations. VeCLIP~\cite{lai2025veclip} and CAPSFUSION~\cite{yu2024capsfusion} inject visual concepts extracted from images into captions with the help of MLLMs, enriching the text content. SynthCLIP~\cite{hammoud2024synthclip} uses text-to-image models to synthesize images and explores fully synthetic CLIP training.

Several works~\cite{zhang2024long, zheng2025dreamlip, wu2024lotlip} focus on releasing the potential of the long text understanding. Long-CLIP~\cite{zhang2024long} reveals that the effective length for CLIP is merely 20 tokens, and fine-tunes the CLIP model by the long captions from ShareGPT4V~\cite{chen2023sharegpt4v}, but it leads to a decline on short text tasks. TULIP~\cite{najdenkoska2024tulip} replaces absolute positional encodings with rotary positional encodings (RoPE) and initializes a new text encoder using model distillation. But the degradation of short-text abilities is severe, recent works (DreamLIP~\cite{zheng2025dreamlip}, LoTLIP~\cite{wu2024lotlip}, and FLAIR~\cite{xiao2024flair}) have to train from scratch on synthetic datasets generated by InstructBLIP~\cite{instructblip}, LLAVA~\cite{liu2024improved} and ShareGPT4V~\cite{chen2023sharegpt4v}. But these works only use a simple prompt ``Describe the image in detail`` to synthesize captions on CC3M~\cite{sharma2018conceptual}, CC12M~\cite{sharma2018conceptual} and YFCC15M~\cite{thomee2016yfcc100m}.



\section{Method}
\label{sec:method}
\begin{figure*}[t] 
	\centering  
	\includegraphics[width=\textwidth]{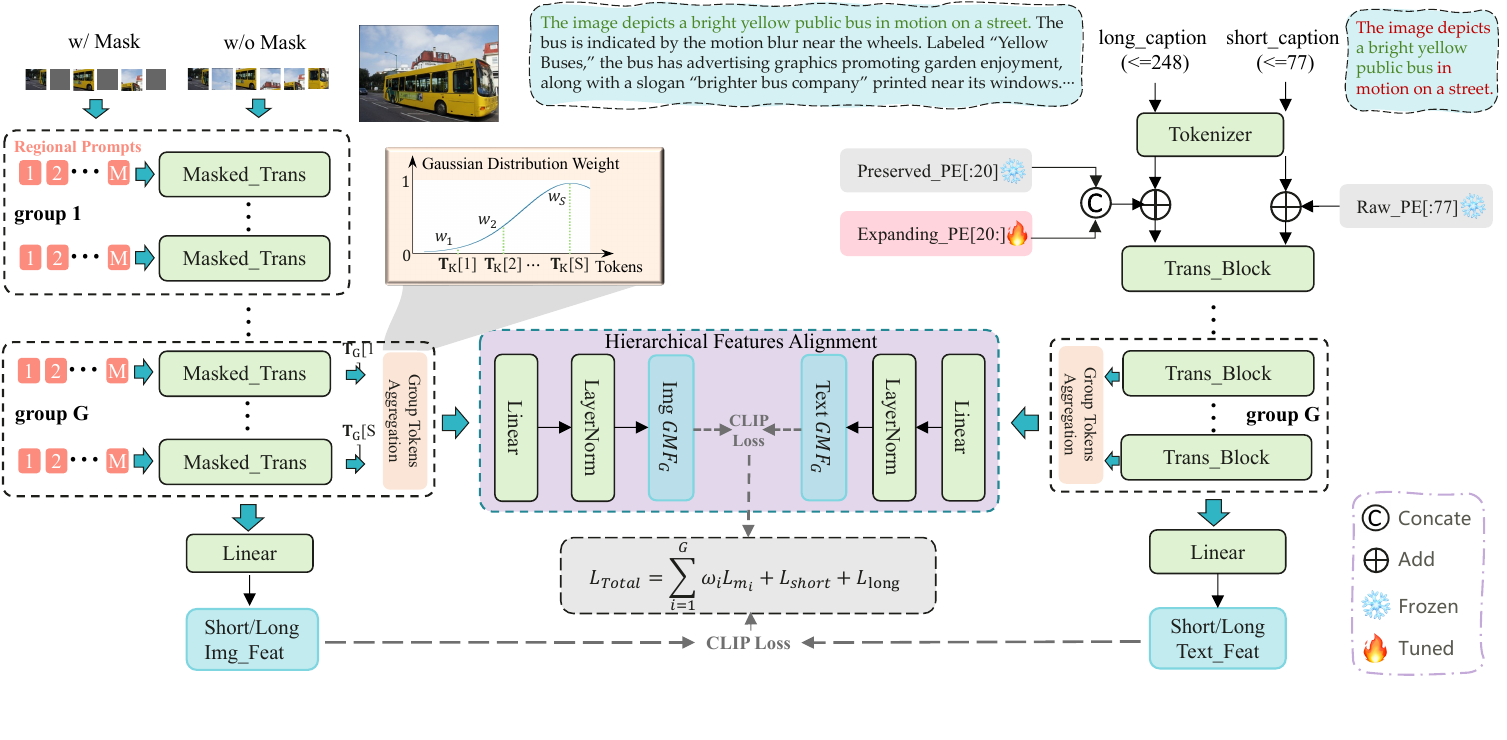}
        \vspace{-3em}
	\caption{Overview of \ours. The image w/o mask aligns with a long caption, while the masked image aligns with a short caption. In the image encoder, regional prompts are employed with the unidirectional mask to extract the regional information. The hierarchical alignment module is designed to associate the middle aggregation features between the image encoder and the text encoder.}
        \vspace{-1em}
	\label{fig:framework}
\end{figure*}

\ours utilizes the incremental training in the synthetic dataset and consists of three components as illustrated in \cref{fig:framework}. The process of long captions synthesis and cleaning is introduced in \cref{sec:3.1}. In \cref{sec:3.2}, we introduce a dual-branch training pipeline. In \cref{sec:3.3}, the regional prompts with unidirectional mask are proposed to extract regional features for fine-grained description. The hierarchical features alignment module proposed in \cref{sec:3.4} aligns the intermediate features in the image encoder and text encoder for contrastive learning.

\subsection{Long-Text Dataset Synthesis and Cleaning} \label{sec:3.1}
We adopt Llama3-LLaVA-NeXT-8b~\cite{liu2024llavanext} to synthesize detailed descriptive long captions. To ensure diversity, we set 20 diverse prompts for synthesis, which are listed in Appendix~\ref{prompts}. The average length of the synthetic captions is around 120 tokens,  which is longer than 18 tokens in the raw captions. We also use Shikra~\cite{chen2023shikra} to synthesize short captions for exploration. We demonstrate the superiority of synthesized short captions over original short captions in \cref{tab:raw-caption} and \cref{tab:raw-caption-2} of the Appendix.

We construct three different scales of synthetic data: (1) \textbf{5M}, including CC3M~\cite{sharma2018conceptual}, VisualGenome~\cite{krishna2017visual}, ShareGPT4V~\cite{chen2023sharegpt4v} and SBU~\cite{ordonez2011im2text}. (2) \textbf{15M}, including \textbf{5M} and CC12M~\cite{sharma2018conceptual}. (3) \textbf{30M}, including \textbf{15M} and YFCC15M~\cite{thomee2016yfcc100m}. Because MLLMs usually bring hallucination information, we removed low-quality captions including repeated words, meaningless sentences, and short results. Some low-quality examples are shown in Appendix~\ref{sec:incorrect caption examples}. The final training data details are shown in \cref{tab:training dataset detail} of the Appendix.


\subsection{Dual-Branch Training Pipeline} \label{sec:3.2}
It is essential to design distinct encoding strategies for texts of varying lengths to enhance the expressiveness of long texts while maintaining the feature extraction capabilities for short texts. To achieve this, we follow Long-CLIP~\cite{zhang2024long} to retain the parameters of the text Transformer blocks from the pre-trained model and modify the position embeddings. We inherit 77 raw position embeddings $PE$ from the pre-trained model and freeze the parameters for short texts. For long texts, we freeze the first 20 position embeddings. Then, we expand the remaining position embeddings (from 21 to 77) through the interpolation method to reach four times of the original length, denoted as:
    \begin{equation} \label{eq:pos_embed}
    \begin{aligned}
        PE_l = Concat(PE[:20],& Intpol(PE[20:],4)), \\
        Intpol(PE, q)[i] =(1&-\lambda) * PE[\lfloor \frac{i}{q} \rfloor] + \\
        \lambda * PE[\lfloor \frac{i}{q} \rfloor +1&], \quad \lambda = \frac{i\%q}{q},
    \end{aligned}
    \end{equation}
where $\lfloor \cdot \rfloor$ defines the floor function. $q$ and $i$ denote the index of the interpolated ratio and the interpolation position, while $\lambda$ represents the assigned weight. Notably, only the positional embedding ($PE$) in Eq.(1) is learnable.

Consequently, the length of the position embeddings for long texts $PE_l$ increases to 248, adequately meeting the requirements in most scenarios. During the training, the parameters of these expanded embeddings are updated to facilitate the extraction of the postpositional information in the text. 

Texts of diverse lengths usually correspond to distinct feature spaces, which require customized image features to match. MAE~\cite{he2022masked} claims that 75\% random masked image retains sufficient semantic information. Therefore, aligning random masked images with short captions is an efficient and low-cost pipeline. Specifically, given the raw image patch embeddings $I \in \mathbb{R}^{N \times D}$, we randomly replace $\alpha \times N$ patch embeddings with learnable parameters initialized by $0$ to denote the masked images $I_m$, where $\alpha$ is the mask ratio and is set as 0.75 at first. Then, we consider the masked image patches $I_m$ and short texts $T_s$ as the pairs for contrastive learning. Conversely, the long texts often include specific details retained in the raw images. Therefore, we take the raw image patches $I$ and long texts $T_l$ as input pairs.


\subsection{Regional Prompts with Unidirectional Mask} \label{sec:3.3}
The [CLS] token interacts with all patch embeddings via the attention mechanism to aggregate the global visual features in the image encoder. However, the capability to recognize local information is insufficient. To address this issue, we introduce several learnable parameters as regional prompts and leverage an unidirectional attention mask to ensure that these prompts attend only to the corresponding regions in the image. Specifically, in the $l$-th Transformer block layer of the image encoder, we interpolate the initial input sequence $([CLS], P_1,\cdots, P_N)$ with $M$ learnable prompts to define our input $([CLS], R^l_1,\cdots, R^l_M, P_1,\cdots, P_N)$, where  $P_i  (i \in [1, N])$ represents the $i$-th patch embedding and $R^l_j (j \in [1, M])$ denotes the $j$-th regional prompt. After the Multi-Head Self-Attention (MHSA) in the $l$-th layer, the input sequences are encoded to $\mathbf{X}^l \in \mathbb{R}^{(1+M+N) \times D}$ where $D$ is the dimension of the channel. Subsequently, each regional prompt $R^l_j$ in $\mathbf{X}^l$ is replaced with a new learnable regional prompt from the next layer $R^{l+1}_j$:
    \begin{equation} \label{eq:prompt_replace}
        \mathbf{X}^l[1:1+M] = (R^{l+1}_1,\cdots,R^{l+1}_M).
    \end{equation}
The procedure above enables each prompt to focus solely on the local features in the current layer, which eliminates the interference of information across different depth layers.

\begin{figure}[t] 
	\centering  
	\includegraphics[width=0.85\linewidth]{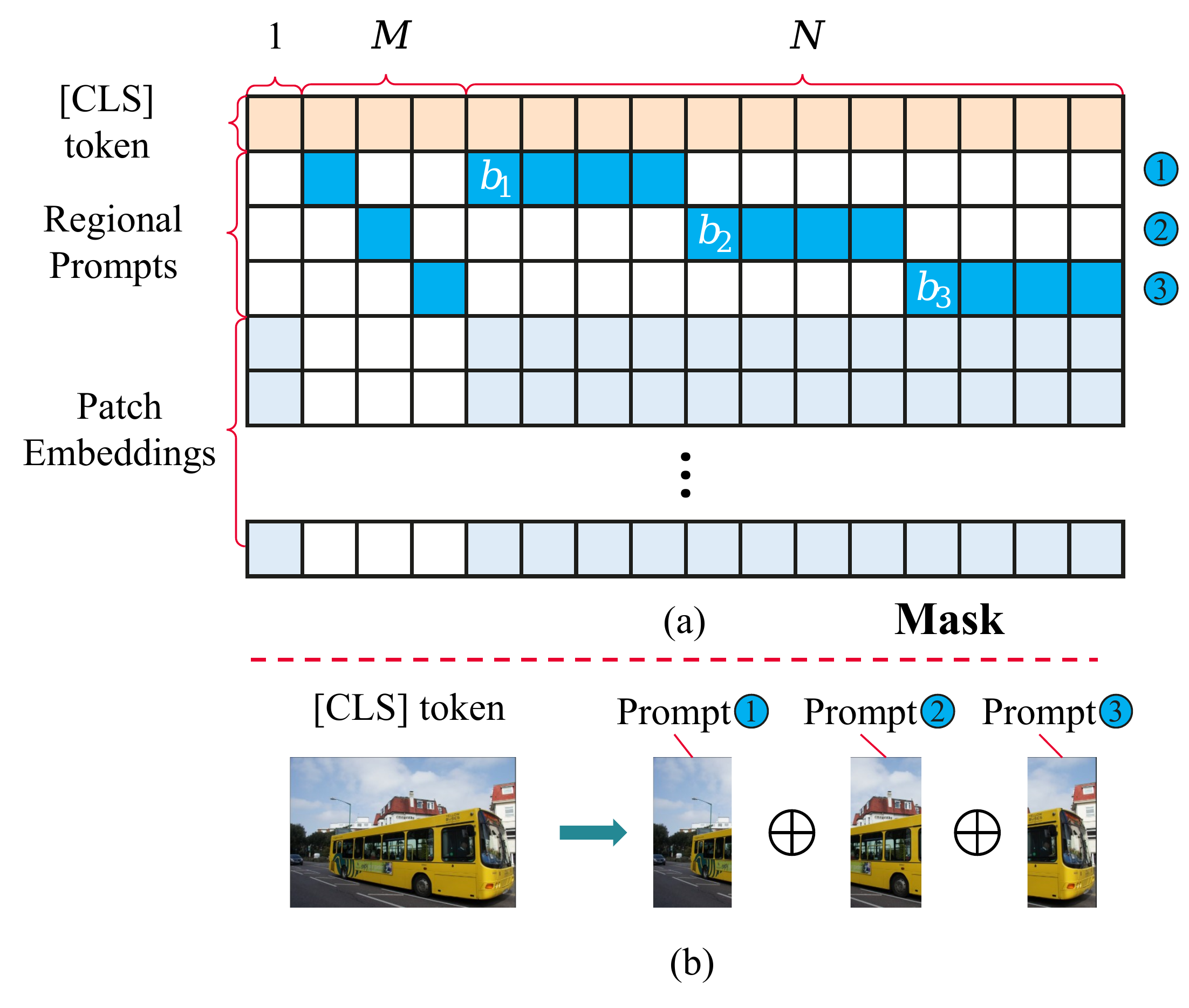}
        \vspace{-1em}
	\caption{Unidirectional mask map is proposed to achieve the unidirectional information propagation from patches to prompts. (a) The illustration of $\mathbf{Mask}$ map. (b) The [CLS] token attends to the global description but each prompt enables to focus on a specific region.}
        \vspace{-1em}
	\label{fig:mask_map}
\end{figure}

During multi-head self-attention, we additionally implement an unidirectional attention mask $\mathbf{Mask}$ to allow the regional prompts to concentrate on the specific local patches while preserving the integrity of the original patch embeddings. As illustrated in \cref{fig:mask_map}, each row represents the mask vector of a query $Q$, which is implemented as follows: the [CLS] token attends to itself as well as all the regional prompts and patch embeddings; the patch embedding $P_i$ focuses on the non-regional prompts partition; each regional prompt $R_j$ attends only to itself and the patch embeddings in the related region, whose mask vector is defined as: 
\begin{equation} \label{eq:region_mask}
    \begin{aligned}
    \mathbf{Mask}[R_j] &= \mathbbm{1}(j,b_j,\cdots,b_j+\lfloor \frac{N}{M} \rfloor -1), \\
    b_j &= 1 + M + j \times \lfloor \frac{N}{M} \rfloor,
    \end{aligned}
\end{equation}
where $\mathbbm{1}(\cdot) \in \mathbb{R}^{1+M+N}$ presents the flag function that the indicated positions are set as $1$ while other places are defined as $0$, and $\lfloor \cdot \rfloor$ is the floor function. $b_j$ denotes the first index of the patches corresponding to the current prompt $R_j$. This method effectively promotes the extraction of local information within regional prompts while restraining the influence of patch embeddings. Then, we multiply the proposed $\mathbf{Mask}$ with the mask map calculated from the self-attention in an element-wise manner to obtain our final attention mask. The $(l+1)$-th Transformer block encoder $ \mathcal{T}^{l+1}_{\text{MHSA}}(\cdot)$ can be formulated as follows: 
\begin{equation} \label{eq:mask_mhsa}
    \begin{aligned}
        \mathbf{X}^{l+1} &= \mathcal{T}^{l+1}_{\text{MHSA}}(\mathbf{X}^l) \\
        &= \text{softmax}(\frac{QK^T}{\sqrt{d}} \odot \mathbf{Mask})V,
    \end{aligned}
\end{equation}
where $\mathbf{X}^l$ and $\mathbf{X}^{l+1}$ are the input and the output of $\mathcal{T}^{l+1}_{\text{MHSA}}(\cdot)$. $Q$, $K$ and $V$ are calculated by multiplying $\mathbf{X}^l$ with the learnable weights $W_Q$, $W_K$ and $W_V$, and $d$ is the channel number of $Q$ and $K$.

\subsection{Hierarchical Feature Alignment} \label{sec:3.4}
Because of the superior complexity of the long text feature spaces, it is not enough to build only the correlation on the vision-language features of the last layer. The intermediate layer features should also exhibit consistency, and this can be achieved via a hierarchical feature alignment module. To be specific, given that there are total $L$ Transformer block layers in the image encoder, $T_l = \mathbf{X}^l[0]$ denotes the [CLS] token in the $l$-th layer. Then, all the $L$ tokens are divided into $G$ groups uniformly with each group containing $S = L/G$ tokens, and the $g$-th group is denoted as $\mathbf{T}_g$.  Then, the Gaussian distribution weights are utilized for Group Tokens Aggregation (GTA) as follows:
\begin{equation} \label{eq: GTA}
        GTA(\mathbf{T}_g) =\sum_{j=1}^{S}  \mathbf{Gaussian}(j;S,1) * \mathbf{T}_g[j].
\end{equation}

Subsequently, the aggregated features $GTA(\mathbf{T}_g)$ will be fed into a linear projection layer, followed by a layer normalization operator to calculate the $g$-th Group Middle Feature (GMF):
\begin{equation}
    GMF_g= LN(Proj(GTA(\mathbf{T}_g))).
\end{equation}

As for the text branch, all of the Transformer blocks are also divided into $G$ groups, followed by the similar strategy above to obtain GMF for the long caption. We also empirically observe that the image features and text features in the shallow layers exhibit larger divergence compared to those in the deeper layers, as shown in \cref{alation:3}. Therefore, we only align the GMF from the $K$-th to the $G$-th group to reduce the computational cost and accelerate model convergence. Furthermore, we utilize the Information Noise Contrastive Estimation (InfoNCE)~\cite{he2020momentum} to calculate the loss $L_{m_i}$ of GMF:
\begin{equation}
\begin{aligned}
        L_{m_i} =& -\sum_{j=1}^B \text{log} \frac{\text{exp}(\text{cos} \langle v_i^j, t_i^j \rangle / \tau)}{\sum_{k=1}^{B}\text{exp}(\text{cos} \langle v_i^j, t_i^k \rangle / \tau)} \\
        &-\sum_{j=1}^B \text{log} \frac{\text{exp}(\text{cos} \langle t_i^j, v_i^j \rangle / \tau)}{\sum_{k=1}^{B}\text{exp}(\text{cos} \langle t_i^j, v_i^k \rangle / \tau)},
\end{aligned}
\end{equation}
where $B$ denotes the batch size, $v_i$ and $t_i$ represent the GMF of the image and text in the $i$-th group. Finally, we multiply each $L_{m_i}$ with weight $\omega_i$ and sum up them with the InfoNCE loss of short-text-image pairs $L_{short}$ and long-text-image pairs $L_{long}$. The final contrastive loss for model training is formulated as:
\begin{equation} \label{eq: loss}
     L = \sum_{i=K}^{G} \omega_i L_{m_i} + L_{short} + L_{long}.
\end{equation}


 \begin{table*}[t]
    \centering
    \resizebox{0.85\linewidth}{!}{
    \begin{tabular}{cccccccccccc}
    \toprule
    \multirow{2}{*}{} &\multirow{2}{*}{Method} &\multirow{2}{*}{Data}
    &\multicolumn{2}{c}{DCI}  &\multicolumn{2}{c}{IIW} & \multicolumn{2}{c}{ShareGPT4V-1k} & \multicolumn{2}{c}{Urban-1k}  &\multirow{2}{*}{Avg.}\\
    \cmidrule(r){4 - 5}
    \cmidrule(r){6 - 7}
    \cmidrule(r){8 - 9}
    \cmidrule(r){10 - 11}
    &&& ~I-to-T~ &  ~T-to-I~ &  ~I-to-T~ &  ~T-to-I~ & ~I-to-T~ &  ~T-to-I~ &  ~I-to-T~ &  ~T-to-I~ &\\
    \midrule
    \multirow{10}{*}{B/16} & CLIP$^\ast$~\cite{radford2021learning} &400M&37.3&34.5&75.2&76.4&78.2&79.6&68.1&53.6&62.8\\
        & Fine-tuned CLIP &1M&46.3&45.4&87.4&85.6&94.1&93.6&80.4&79.8&76.6\\
        & Long-CLIP~\cite{zhang2024long} &1M& 51.1&57.0&89.2&86.9&94.6& 93.3 & 78.9&79.5&76.8\\
        & \ours &1M& \textbf{59.7} & \textbf{63.0} & \textbf{93.8} & \textbf{95.6}& \textbf{95.5 } & \textbf{ 94.1 } & \textbf{ 80.9 } & \textbf{ 81.1} & \textbf{82.6}\\
        \cmidrule(){2 - 12}
    &SigLIP$^\ast$~\cite{zhai2023sigmoid}&12B&57.8&56.2&91.9&91.0&85.8&83.4&  62.7 & 62.1 & 73.9 \\
        &FLAIR$^\ast$~\cite{xiao2024flair}&30M&61.3&66.2&-&-&98.5&98.0& 83.6 & 87.7 & - \\
        & LoTLIP$^\ast$~\cite{wu2024lotlip}&100M& 62.1&61.0&93.9&92.5&96.5& 95.5 & 77.8 & 76.5 & 81.9 \\
        & \ours&5M & 67.1 & 67.5 & 96.9 & 96.7&98.1 & 97.9 & 88.0 & 90.8 & 87.8\\
        & \ours&15M &69.2&69.9&97.1&97.2&98.3&98.2&88.0&93.7&88.9\\
        & \ours &30M& \textbf{70.7} & \textbf{70.7} & \textbf{97.4} & \textbf{97.4} & \textbf{98.6}&\textbf{98.5} & \textbf{90.8} & \textbf{94.6} & \textbf{89.8}\\
    \midrule
    \multirow{7}{*}{L/14} & CLIP$^\ast$~\cite{radford2021learning} &400M& 37.9&35.9&78.6&80.2&81.8&84.0&68.7&52.8&65.0\\
    & Fine-tuned CLIP &1M&46.9&46.2&88.6&88.5&95.3&95.4&78.0&76.5&76.9\\
    & Long-CLIP~\cite{zhang2024long} &1M&51.7&57.4&91.2&90.1&95.8&95.6&82.7&86.1&79.2\\
        & \ours &1M&  \textbf{65.1} & \textbf{66.7} & \textbf{96.2} & \textbf{97.1} &\textbf{98.1} & \textbf{98.6} &  \textbf{86.8} & \textbf{87.7} &\textbf{85.8}\\
        \cmidrule(){2 - 12}
        & \ours&5M&68.1 & 69.9 & 97.1 & 97.2&98.5 & 98.0 & 88.1 & 93.0 & 88.7\\
        & \ours&15M& 69.2&70.2&97.7&98.1&98.7&98.1&89.5&94.3&89.5\\
        & \ours &30M& \textbf{72.0} & \textbf{74.2} & \textbf{97.9} & \textbf{98.2} & \textbf{99.0} & \textbf{98.3} & \textbf{93.7} & \textbf{96.3}&\bfseries{91.2}\\
    \bottomrule
    \end{tabular}
    }
    \caption{Zero-shot long text-image retrieval benchmarks. I2T and T2I indicate the R@1 score on image-to-text and text-to-image retrieval. SigLIP~\cite{zhai2023sigmoid}, LoTLIP~\cite{wu2024lotlip} and FLAIR(30M)~\cite{xiao2024flair} do not have L/14 results. $^\ast$models are trained from scratch. The best results are \textbf{bold}.}
    \vspace{-1em}
    \label{tab:long-retrieval}
\end{table*}
\section{Experiments}

\subsection{Experimental Setup}


\textbf{Downstream datasets.} 
To evaluate the effectiveness of our model, we select three zero-shot tasks following~\cite{zhang2024long}: short-text-image retrieval, long-text-image retrieval, and image classification. For long-text-image retrieval, following LoTLIP~\cite{wu2024lotlip} and Long-CLIP~\cite{zhang2024long}, we evaluate method on datasets with long captions, including ShareGPT4V-1k~\cite{chen2023sharegpt4v}, Urban-1k~\cite{zhang2024long}, DCI~\cite{urbanek2024picture}, and IIW~\cite{garg2024imageinwords} and report the Recall at 1 (R@1) metric. In DCI~\cite{urbanek2024picture} and IIW~\cite{garg2024imageinwords}, all images with human-authored long captions are used for evaluation. For short-text-image retrieval, we use the 5k validation set of COCO~\cite{chen2015cococaption} and 1k test set of Flickr30k~\cite{plummer2015flickr30k} for evaluation and present the Recall at 1, 5 and 10 (R@1, R@5 and R@10). For image classification, we evaluate on ImageNet-1K~\cite{deng2009imagenet}, ImageNet-V2~\cite{recht2019imagenet}, ImageNet-O~\cite{hendrycks2021natural}, ImageNet-A~\cite{hendrycks2021natural}, CIFAR-10~\cite{2009Learning} and CIFAR-100~\cite{2009Learning} and report the top-1 Accuracy (Acc@1). 

\noindent\textbf{Training setup.} 
For a fair comparison, our experiment setup follows Long-CLIP~\cite{{zhang2024long}}. For results without specifically indicating data scales, the training dataset is ShareGPT4V~\cite{chen2023sharegpt4v}, which contains 1M long-text-image pairs. For results with data scales, the training datasets are our synthetic data. Two variants of Vision Transformer are used as the image encoder in our experiments, \ie ViT-B/16 and ViT-L/14, while the text encoder is a vanilla Transformer. The image size is 224 $\times$ 224, and the input text sequence length is truncated or padded to 248. We train the model on 16 $\times$ A800 GPUs with a batch size of 2048. The other hyperparameters are under the same setting as Long-CLIP~\cite{zhang2024long} (\eg, learning rate, warmup steps, and weight decay). Detailed training settings are shown in Appendix \ref{sec:train_hyper}.

\subsection{Comparison with Previous Methods}
In this section, our method is trained on ShareGPT4V~\cite{chen2023sharegpt4v} (1M) and tested on numerous open-vocabulary benchmarks, including zero-shot retrieval and classification. We compare \ours against the state-of-the-art approaches to prove the effectiveness of our method. 

\noindent\textbf{Long text-image retrieval.} The results in \cref{tab:long-retrieval} demonstrate that \ours has superior ability in long-text understanding. Comparing models trained on ShareGPT4V(1M), \ours surpasses Long-CLIP in the long text-image retrieval task, obtaining higher R@1 scores on both DCI (I2T: +8.6\%, T2I: +6\%) and IIW (I2T: +4.6\%, T2I: +8.7\%) datasets. The average improvements with ViT-B/16 and ViT-L/14 image encoders can even achieve 5.8\% and 7.3\% compared to Long-CLIP.

\noindent\textbf{Short text-image retrieval.}
\cref{tab:short-retrieval} shows the main results of short-text-image retrieval in COCO~\cite{chen2015cococaption} and Flickr30k~\cite{plummer2015flickr30k} datasets. With the B/16 encoder, \ours outperforms Long-CLIP~\cite{zhang2024long} in the text-to-image retrieval task, obtaining higher R@1 scores on COCO (+4.4\%), and Flickr30k (T2I: +5.1\%) datasets. \ours surpasses Long-CLIP with an average 2.2\% improvement with the ViT-L/14 image encoder in the R@1 metric. \ours also outperforms TULIP~\cite{najdenkoska2024tulip} which uses two-stage training with distillation and fine-tuning on 1M data. The above results show that \ours can enhance the long-text understanding while maintaining the generalization capability on short-text tasks.

\noindent\textbf{Zero-shot classification tasks.} As shown in \cref{tab:classification}, \ours also achieves promising performance. In particular, \ours attains a remarkable improvement on two challenging adversarial out-of-distribution datasets, ImageNet-O~\cite{hendrycks2021natural} and ImageNet-A~\cite{hendrycks2021natural}.  \ours provides robustness and generalization capabilities of \ours in handling complex and adversarial scenarios.

\subsection{Scalability Analysis}
Due to the degradation of short-text capabilities, recent works have to train from scratch on reconstructed datasets, incurring high resource costs. Our method employs incremental training and aligns with CLIP's original short-text feature space better. 

We further inspect the scalability of \ours across three data scales: 5M, 15M, and 30M. Our investigation in \cref{tab:long-retrieval} reveals that synthetic long-text captions exhibit remarkable scalability. When including SOTA models, \ours trained on 5M synthetic data and B/16 image encoder outperforms SigLIP, FLAIR, and LoTLIP on all pre-training datasets by a large margin. When we move to larger datasets with 30M synthetic data, \ours surpasses the previous SOTA in the long text-image retrieval task, obtaining higher R@1 scores on DCI (I2T: +8.6\%, T2I: +9.7\%), IIW (I2T: +3.5\%, T2I: +4.9\%), and Urban-1k (I2T: +7.2\%, T2I: +6.9\%) datasets.

Models like EVA-CLIP and FLIP are pre-trained on large short-text datasets, with EVA-CLIP even trained at a 2 billion scale. This leads to significant degradation of short-text capabilities in prior long-text understanding works. Benefiting from our training pipeline, CLIP's original short-text abilities are preserved and continuously enhanced with increased training data. As shown in \cref{tab:short-retrieval}, with 30M training data, \ours outperforms the previous work in the text-to-image retrieval task, obtaining higher R@1 scores on COCO (+6.9\%), and Flickr30k (T2I: +8.4\%) datasets.

In summary,  \ours outperforms state-of-the-art approaches by 13\% and 5\% on long-text and short-text benchmarks, respectively.


\begin{table*}[t]
    \centering
    \resizebox{\textwidth}{!}{
    \begin{tabular}{c c c c  c  c  c c c c c c c c c c}
    \toprule
    \multirow{3}{*}{} &\multirow{3}{*}{Method}&\multirow{3}{*}{Data}
     & \multicolumn{6}{c}{COCO} & \multicolumn{6}{c}{Flickr30k} & \multirow{2}{*}{Avg.} \\
    &&& \multicolumn{3}{c}{Image-to-Text} &  \multicolumn{3}{c}{Text-to-Image}&  \multicolumn{3}{c}{Image-to-Text}&  \multicolumn{3}{c}{Text-to-Image}&\\
    \cmidrule(r){4 - 6}
    \cmidrule(r){7 - 9}
    \cmidrule(r){10 - 12}
    \cmidrule(r){13 - 15}
    &&& R@1~ & R@5~ &R@10~ & R@1~ & R@5~ &R@10~ & R@1~ & R@5~ &R@10~ & R@1~ & R@5~ &R@10~ & R@1\\
    \midrule
    \multirow{10}{*}{B/16} & CLIP$^\ast$~\cite{radford2021learning} &400M&51.8 &76.8 &84.3 &32.7 &57.7&68.2& 82.2 & 96.6 & 98.8 & 62.1& 85.7 & 91.8 & 57.2\\
        & Long-CLIP~\cite{zhang2024long} &1M&57.6 & 81.1& 87.8& 40.4& 65.8& 75.2& 87.9 & 97.2& 98.9 & 72.3 & 92.2 & 95.6&64.6\\ 
        & TULIP~\cite{najdenkoska2024tulip}&1M&56.8&80.3&-&40.7&66.1&-&86.9&96.4&-&73.7&93.6&-&64.5\\ 
        & \ours &1M& \textbf{ 60.9 } & \textbf{ 83.4 } & \textbf{ 90.2 } & \textbf{ 44.8 } & \textbf{ 70.2 } & \textbf{ 79.5 } & \textbf{ 88.4 } & \textbf{ 98.5 } & \textbf{ 99.5 } & \textbf{ 77.4 } & \textbf{ 94.8 } & \textbf{ 97.1 } & \textbf{67.9}\\
        \cmidrule(){2 - 16}
        & EVA-CLIP$^\ast$~\cite{sun2023eva} &2B&58.7& 80.7& 88.2 & 42.2 & 66.9 & 76.3 & 85.7 & 96.7 & 98.9 & 71.2 & 91.0 & 94.7& 64.5\\ 
        & LoTLIP$^\ast$~\cite{wu2024lotlip} &30M&59.7& 81.5& -- & 38.1 & 63.8& -- & 86.9 & 97.8 & -- & 65.2 & 88.0 & --& 62.5\\ 
        & DreamLIP$^\ast$~\cite{zheng2025dreamlip}&30M&58.3& 81.6& 88.8 & 41.1 & 67.0 & 76.6 & 87.2 & 97.5 & 98.8 & 66.4 & 88.3 & 93.3 & 63.3\\ 
        & \ours &5M&61.3 & 84.9& 91.2& 47.0& 72.4& 81.4& 89.9 & 98.8& 99.7 & 78.4 & \textbf{95.2} & \textbf{97.7} &69.2\\ 
        & \ours&15M&61.2&84.7&\textbf{91.8}&48.7&\textbf{74.3}&\textbf{82.7}&89.1&98.4&99.7&79.5&95.1&97.6&69.6\\ 
        & \ours &30M& \textbf{62.3} & \textbf{85.4} & 91.4 & \textbf{49.1} & 73.8 & 82.4 & \textbf{90.5} & \textbf{99.0} & \textbf{99.8} & \textbf{79.6} & 94.9 & 97.4&\textbf{70.4}\\
    \midrule
    \multirow{9}{*}{L/14} & CLIP$^\ast$~\cite{radford2021learning} &400M& 56.3 & 79.3 & 86.7 & 36.5 & 61.0 & 71.1 & 85.2 & 97.3 & 99.0 & 65.2 & 87.3 & 92.0 & 60.8\\
    & Long-CLIP~\cite{zhang2024long} &1M& 58.3 & 81.4& 88.2 & 45.1 & 70.4& 79.3 & 90.9 & 98.8 & 99.5 & 78.7 & 94.5 & 97.1 & 68.3\\
    & TULIP~\cite{najdenkoska2024tulip}&1M&62.6&84.7&-&46.1&71.1&-&92.3&99.3&-&79.0&94.8&-&70.0\\ 
    & \ours&1M&\textbf{63.4}&\textbf{85.8}&\textbf{91.4}&\textbf{46.5}&\textbf{72.0}&\textbf{80.7}&\textbf{93.0}&\textbf{99.5}&\textbf{99.6}&\textbf{79.2}&\textbf{95.9}&\textbf{97.4}&\textbf{70.5}\\
    \cmidrule(){2 - 16}
        & FLIP$^\ast$~\cite{li2023scaling} &400M& 60.2 & 82.6 & 89.9 & 44.2 & 69.2 & 78.4 & 89.1 & 98.5 & 99.6 & 75.4 & 92.5 & 95.9 & 67.2\\
        & EVA-CLIP$^\ast$~\cite{sun2023eva} &2B& 63.7 & 84.3 & 90.4 & 47.5 & 71.2& 79.7 & 89.7 & 98.6 & 99.2 & 77.3 & 93.6 & 96.8&69.6 \\
        & \ours&5M&63.2&85.8&91.5&50.5&75.4&83.6&\textbf{92.5}&99.1&99.9&82.5&96.6&98.2&72.1\\ 
        & \ours&15M&63.7&\textbf{86.8}&\textbf{92.1}&51.9&76.2&84.4&90.7&99.3&99.9&83.8&96.6&98.3&72.5\\ 
        & \ours&30M& \textbf{64.5} & 86.5 & 91.9 & \textbf{52.6} & \textbf{77.2} & \textbf{84.9} & 91.5 & \textbf{99.8} & \textbf{99.9} & \textbf{84.1} & \textbf{96.7} & \textbf{98.4}&\textbf{73.2}\\
    \bottomrule
    \end{tabular}
    }
    \caption{Results of zero-shot short text-image retrieval on the COCO~\cite{chen2015cococaption} validation set and the 1k Flickr30K~\cite{plummer2015flickr30k} test set. LoTLIP~\cite{wu2024lotlip} and DreamLIP~\cite{zheng2025dreamlip} do not provide L/14 results. FLIP~\cite{li2023scaling} does not provide B/16 results. $^\ast$models are trained from scratch. The best results are \textbf{bold}.}
    \vspace{-2em}
    \label{tab:short-retrieval}
\end{table*}

\begin{table}[t]
    \centering 
    \tabcolsep=5pt
    \resizebox{\linewidth}{!}{
    \begin{tabular}{c c c cc c c c c}
    \toprule
    &Method&\rotatebox{90}{IN-1k} &\rotatebox{90}{IN-O} &\rotatebox{90}{IN-A}& \rotatebox{90}{IN-V2} &\rotatebox{90}{Cifar10}&\rotatebox{90}{Cifar100}&\rotatebox{90}{Average}\\
    \midrule
    \multirow{4}{*}{B/16} & CLIP~\cite{radford2021learning} & \textbf{68.4} & 42.2 &38.4& \textbf{61.9} &\underline{90.8} &67.3 & 61.5\\
        &Fine-tuned CLIP & 55.1 & 31.7 &30.5& 44.8 & 83.9 & 59.2 & 50.9 \\
        & Long-CLIP~\cite{zhang2024long} & 66.8 & \underline{42.7} &\underline{46.0}& 61.2 & 90.7 & \underline{69.3} & \underline{62.7}\\
        & \ours&\underline{68.0}&\textbf{44.1}&\textbf{49.8}&\underline{61.8}&\textbf{91.9}&\textbf{70.6}&\textbf{64.4}\\
    \midrule
    \multirow{4}{*}{L/14} & CLIP~\cite{radford2021learning} & \textbf{75.5} & 31.9 &46.4& \textbf{69.9} & \underline{95.5} & 76.8 & 66.0 \\
        & Fine-tuned CLIP & 58.4 & 29.2 &35.8& 52.7 & 92.7 & 68.7 & 56.3\\
        & Long-CLIP~\cite{zhang2024long}& 73.5 & \underline{33.7} &\underline{61.0} & 67.9 & 95.3 & \underline{78.5} & \underline{68.3}\\
        & \ours&\underline{73.7}&\textbf{35.9}&\textbf{66.7}&\underline{68.8}&\textbf{96.2}&\textbf{78.9}&\textbf{71.4}\\
    \bottomrule
    \end{tabular}
    }
    \caption{Top-1 accuracy for zero-shot classification on: ImageNet-1K~\cite{deng2009imagenet}, ImageNet-O~\cite{hendrycks2021natural}, ImageNet-A~\cite{hendrycks2021natural}, ImageNet-V2~\cite{recht2019imagenet}, CIFAR-10~\cite{2009Learning} and CIFAR-100~\cite{2009Learning}. The best and second-best results are \textbf{bold} and \underline{underlined}. }
    \vspace{-1.5em}
    \label{tab:classification}
\end{table}

\subsection{Ablation Study} \label{sec:4.4}
\noindent\textbf{Model Components.} To assess the effectiveness of the proposed modules, we conduct the ablation studies through incremental training on the ShareGPT4V~\cite{chen2023sharegpt4v} dataset. The image encoder is ViT-L/14~\cite{alexey2020image} and the text encoder is the same as~\cite{radford2021learning}. In \cref{tab:main_ablation}, we analyze the components of \ours: dual-branch training pipeline (DB), hierarchical feature (HF) alignment, regional prompts (RP), and unidirectional mask (UM).

Long-CLIP~\cite{zhang2024long} is the baseline of the ablation (0). Changing the training pipeline to the dual-branch (DB) type leads to performance improvements across all metrics (1), achieving an 8.8\%/4\% boost in R\@1 for DCI long text-image retrieval, which demonstrates its contribution to long-text understanding. Hierarchical feature (HF) alignment also provides decent gain for all benchmarks (2). Adding regional prompts (RP) improves the performance in each task (3 and 4). The interpolation of regional prompts (RP) further improves the performance of \ours in various metrics and even achieves the best performance on COCO's T2I task (5 and 6). Additionally, the unidirectional mask (UM) alleviates the degradation of generalization capability in short texts, achieving 0.9\% improvement in image-to-text retrieval on the COCO~\cite{chen2015cococaption} dataset. \ours with all components achieves the best performance (7). Overall, the dual-branch training pipeline is the foundation of \ours, giving the ability to understand long text, while other components contribute to continued performance growth.

\begin{table}[t]
	\centering\scriptsize
	\SetTblrInner{rowsep=1pt}
	\SetTblrInner{colsep=4pt}
	\resizebox{\columnwidth}{!}{
		\begin{tblr}{
			cells={halign=c,valign=m},   %
			hline{1,11}={0.7pt},       %
					hline{3}={},       %
					hline{2}={2-5}{leftpos = -1, rightpos = -1, endpos},
					hline{2}={6-7}{leftpos = -1, rightpos = -1, endpos},
					hline{2}={8-9}{leftpos = -1, rightpos = -1, endpos},
					hline{2}={10-11}{leftpos = -1, rightpos = -1, endpos},
					cell{1}{2}={c=4}{},
					cell{1}{6}={c=2}{},
					cell{1}{8}={c=2}{},
					cell{1}{10}={c=2}{},
				}
			  & Method      &             &             &             & DCI &          & IIW &          & COCO &           \\
			& \textbf{DB} & \textbf{HF} & \textbf{RP} & \textbf{UM} & I2T          & T2I    & I2T           & T2I    & I2T              & T2I     \\

                0 &   &             &             &             & 51.7           & 67.4     & 91.2            & 90.1     & 58.3              & 45.1    \\
			1 & \checkmark  &             &             &             & 60.5           & 61.4     & 94.0            & 95.2     & 60.9              & 45.9    \\
			2 &            & \checkmark  &             &             & 53.3           & 58.5     & 91.9            & 92.3     & 59.6               & 45.3    \\
			3 & \checkmark  &             & \checkmark  &             & 62.4           & 62.6     & 94.5           & 95.6     & 61.7               & 46.1     \\
			4 &            & \checkmark  & \checkmark  &             & 54.5           & 59.8     & 92.8           & 92.7     & 60.3               & 45.6     \\
                5 &    \checkmark        &  & \checkmark  &     \checkmark        &     63.5     &  63.1   &     95.9       &  96.1    &       63.0       & \bf 46.8   \\
			6 &            & \checkmark  & \checkmark  & \checkmark  & 56.3           & 62.7     & 93.5            & 93.7     & 61.2              & 45.9   \\
			7 & \checkmark  & \checkmark  & \checkmark  & \checkmark  & \bf 65.1       & \bf 66.7 & \bf 96.2        & \bf 97.1 & \bf63.4           & 46.5 \\
		\end{tblr}}
	\caption{Ablation study on different components of \ours. \textbf{DB}: Dual-Branch training pipeline, \textbf{HF}: Hierarchical Feature alignment, \textbf{RP}: Regional Prompts, \textbf{UM}: Unidirectional Mask.
    }
        \vspace{-2em}
	\label{tab:main_ablation}
\end{table}

\noindent\textbf{Ablation on different input schemes.} In the default implementation, the same position embedding is used for short and long texts, and only the original image patches are utilized for feature extraction. When we modify the position embedding strategies to accommodate texts of varying lengths, the improvement can be observed in the retrieval task, as shown at the top of \cref{tab:inputs_way}. Furthermore, aligning the masked image features with short caption features results in higher recall. We also find that preserving the original length of image patches by replacing the masked patch embeddings with learnable parameters yields better performance. As shown at the bottom of \cref{tab:inputs_way}, although discarding the masked patches reduces the computational cost and memory occupancy, the recall significantly drops compared to the strategy that preserves the length.

\begin{table}[t]
    \centering 
    \resizebox{\linewidth}{!}{
    \tabcolsep=5pt
    \begin{tabular}{c c| c c| c cc c c c}
    \toprule
   \multirow{2}{*}{Init.} & \multirow{2}{*}{Pos.} & \multirow{2}{*}{Pre.} & \multirow{2}{*}{Dis.} & \multicolumn{2}{c}{DCI} & \multicolumn{2}{c}{COCO} & \multirow{2}{*}{Mem./G} & \multirow{2}{*}{Time/ms}\\
    & & & & I2T & T2I& I2T & T2I & & \\
    \midrule
    \usym{2713}& \usym{2717}&\usym{2717} & \usym{2717}& 62.6 & 62.9 & 61.2 & 45.9 & -- & -- \\
    \usym{2717} & \usym{2713}& \usym{2717} & \usym{2717}& \textbf{64.3} & \textbf{64.2} & \textbf{62.1} & \textbf{46.1} & -- & -- \\
    \midrule
   \usym{2717} & \usym{2713}& \usym{2713}&\usym{2717} &\textbf{65.1}&\textbf{66.7}&  \textbf{63.4} & \textbf{46.5} & 17.2 & 61.92\\
    \usym{2717}& \usym{2713}&\usym{2717} & \usym{2713}& 62.7 & 63.4 & 62.0 & 45.3 & 16.3 & 58.32\\
    \bottomrule
    \end{tabular}
    }
     \vspace{-1em}
    \caption{Ablation on different inputs schemes. ``Init.'' means the configuration of position embedding follows Long-CLIP \cite{zhang2024long}. ``Pos.'' means conducting the different position embedding for texts. ``Pre.'' means preserving the masked image patches, and ``Dis.'' means discarding the masked image patches. ``Mem./G'' and ``Time/ms'' are memory occupancy and time cost on a A800 GPU.}
    \vspace{-1em}
    \label{tab:inputs_way}
\end{table}

\begin{table}[t]
    \centering 
    \scriptsize
    \resizebox{0.7\linewidth}{!}{
    \begin{tabular}{cc c cc c  c}
    \toprule
   \multirow{2}{*}{UM} & \multirow{2}{*}{Num.}& \multicolumn{2}{c}{DCI} & \multicolumn{2}{c}{COCO} \\
    & &  I2T & T2I& I2T & T2I \\
    \midrule
    \usym{2717}& 0 & 60.9 & 62.7 & 62.7 & 46.4 \\
    \usym{2717} & 4 & 62.5 & 64.0 & 62.9 & 46.2 \\
    \usym{2713} & 1 & 62.7 & 64.3 & \textbf{63.6} & 46.3 \\
    \usym{2713}& 2 & 64.2 & 64.5 & 63.1 & 46.4  \\
    \usym{2713}& 4 & \textbf{65.1} & \textbf{66.7} & 63.4 & \textbf{46.5}\\
    \usym{2713}& 8 & 64.8 & 65.4 & 63.0 & 46.5  \\
    \bottomrule
    \end{tabular}
    }
    \vspace{-1em}
    \caption{Ablation on the number M of regional prompts. ``UM'' means utilizing the unidirectional mask. ``Num.'' means the number of regional prompts.}
    \vspace{-2em}
    \label{tab:regional prompts}
\end{table}

\noindent\textbf{Ablation on the number of regional prompts.} The ablation study in \cref{tab:regional prompts} investigates the optimal number of regional prompts, which is represented by $M$ in \cref{sec:3.3}. We interpolate different numbers of prompts in the image encoder. A larger number of regional prompts allows each prompt to attend to a smaller region, enabling finer-grained information capture, as described in \cref{eq:region_mask}. Interestingly, when the number of regional prompts is 1, the image-text retrieval performance on COCO is the highest, demonstrating that more regional prompts aid in extracting local features, while fewer prompts benefit short-text image-text retrieval. When the number of prompts is set to 4, our approach achieves the best performance on average. 

\noindent\textbf{Ablation on different groups of hierarchical feature alignment.} \label{alation:3} As a more reasonable contrastive learning strategy, hierarchical alignment also demonstrates its effectiveness in long-text tasks. We divide all the Transformer blocks into 6 groups both in the image encoder and the text encoder. As illustrated in \cref{tab:hier_contrast}, applying the hierarchical contrastive learning from the $4$-th group to the $6$-th group achieves 1.6\%/3.6\% improvement on the DCI benchmark. The last column shows that the GMF loss steadily decreases as the group depth increases. Moreover, the weight of each GMF loss should increase incrementally, as deeper features are more critical for alignment. Finally, we set the weights for GMF loss as $0.2$, $0.4$, and $0.8$ for $4$-th, $5$-th, and $6$-th groups, respectively.

\noindent\textbf{Other Ablations.} 
It should be noted that we employ long position embeddings across all downstream tasks when referring. The performance comparison of different position embeddings is shown in \cref{tab:suppl_ablation} of the Appendix. \cref{tab:suppl_ablation} also presents more ablations on the efficacy of region prompts and masks.

\begin{table}[t]
    \scriptsize
    \centering 
    \resizebox{0.9\linewidth}{!}{
    \begin{tabular}{c c c cc c  c}
    \toprule
   \multirow{2}{*}{Hier} & \multirow{2}{*}{Groups} & \multicolumn{2}{c}{DCI} & \multicolumn{2}{c}{COCO} & GMF Loss\\
    & & I2T & T2I& I2T & T2I & Avg.\\
    \midrule
    \textit{w/o} & - &63.5&  63.1 & 63.1 & 46.5 & - \\
    \textit{w/}& $[1,\quad 6]$ &  64.0 & 63.9 & 63.1 & 46.3 & 4.48 \\
    \textit{w/}& $[3,\quad6]$  & 64.8 & 65.4 & 62.8 & 46.1 & 3.49 \\
    \textit{w/}& $[4,\quad6]$  & \textbf{65.1} & \textbf{66.7}  & \textbf{63.4} &  \textbf{46.5} & 2.65 \\
    \textit{w/}& $[5,\quad6]$  & 64.7 & 65.6 & 62.7  & 46.2 & 1.71\\
    \bottomrule
    \end{tabular}
    }
    \caption{Ablation on the hierarchical contrastive learning with different groups included.}
    \vspace{-1em}
    \label{tab:hier_contrast}
\end{table}

\begin{figure}[t]
	\centering
	\includegraphics[width=0.9\linewidth]{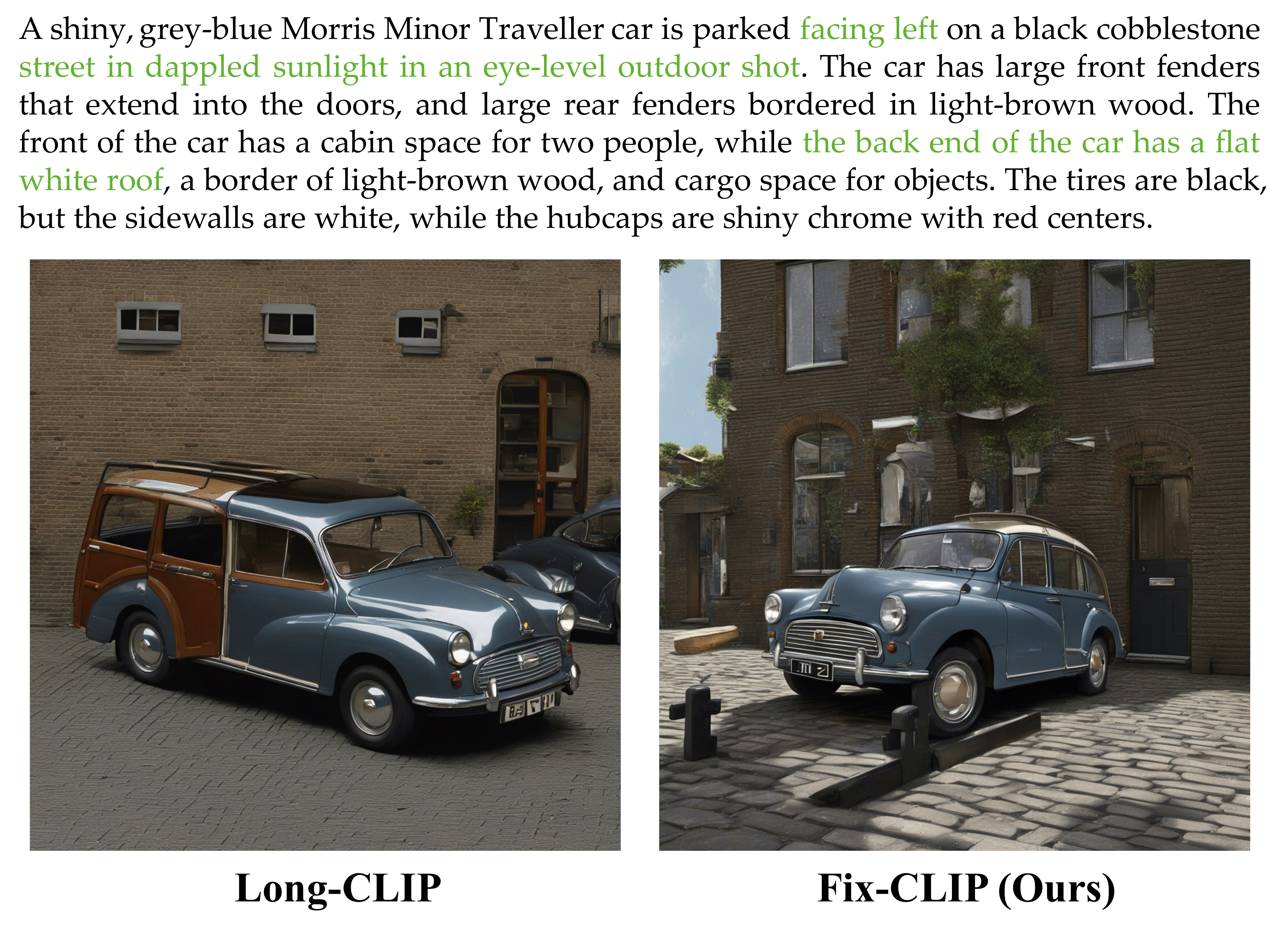}
        \vspace{-1em}
	\caption{Comparison on the text-to-image generation performance. We replace the text encoder in the stable diffusion model with our or Long-CLIP's text encoder.}
        \vspace{-1em}
	\label{fig:SD}
\end{figure}

\subsection{Text-to-Image Generation}
\ours can be integrated into Stable-Diffusion-XL for text-to-image generation in a plug-and-play manner. We replace the CLIP-L text encoder with \ours-L. Benefiting from the effective modules, \ours outperforms Long-CLIP~\cite{zhang2024long} in understanding long texts. As demonstrated in \cref{fig:SD}, images generated by \ours better represent detailed information in long texts, such as object orientation, material, color, background, and interaction details. More generated images are provided in Appendix \ref{sec: more generation examples}.



\section{Conclusion}
In this work, we propose \ours to improve the long-text understanding capability while preserving the short-text ability of CLIP. Considering the distinct feature spaces between short and long texts, we design a dual-branch training pipeline to align short and long texts with masked and raw images, respectively. Then, the learnable regional prompts with unidirectional masks are proposed to extract the local features from patch embeddings. We employ a hierarchical alignment module to establish more precise correspondence between intermediate-level textual and visual representations. To explore the performance limits of our model, we leverage MLLMs to synthesize long captions from 30M images and clean the data for training. \ours outperforms prior works on numerous open-vocabulary tasks across various training data scales and serves as an effective backbone for diffusion models.
{
    \small
    \bibliographystyle{ieeenat_fullname}
    \bibliography{clip}

\begin{thebibliography}{67}
\providecommand{\natexlab}[1]{#1}
\providecommand{\url}[1]{\texttt{#1}}
\expandafter\ifx\csname urlstyle\endcsname\relax
  \providecommand{\doi}[1]{doi: #1}\else
  \providecommand{\doi}{doi: \begingroup \urlstyle{rm}\Url}\fi

\bibitem[Abdollah et~al.(2024)Abdollah, Izadi, Saghafian, Vahidimajd, Mozafari,
  Mirzaei, Samiei, and Baghshah]{abdollah2024comalign}
Ali Abdollah, Amirmohammad Izadi, Armin Saghafian, Reza Vahidimajd, Mohammad
  Mozafari, Amirreza Mirzaei, Mohammadmahdi Samiei, and Mahdieh~Soleymani
  Baghshah.
\newblock Comalign: Compositional alignment in vision-language models.
\newblock \emph{arXiv preprint arXiv:2409.08206}, 2024.

\bibitem[Alexey(2020)]{alexey2020image}
Dosovitskiy Alexey.
\newblock An image is worth 16x16 words: Transformers for image recognition at
  scale.
\newblock \emph{arXiv preprint arXiv: 2010.11929}, 2020.

\bibitem[Cai et~al.(2024)Cai, Cao, Chen, Chen, Chen, Chen, Chen, Chen, Chen,
  Chu, et~al.]{cai2024internlm2}
Zheng Cai, Maosong Cao, Haojiong Chen, Kai Chen, Keyu Chen, Xin Chen, Xun Chen,
  Zehui Chen, Zhi Chen, Pei Chu, et~al.
\newblock Internlm2 technical report.
\newblock \emph{arXiv preprint arXiv:2403.17297}, 2024.

\bibitem[Chen et~al.(2023{\natexlab{a}})Chen, Yu, Ge, Yao, Xie, Wu, Wang, Kwok,
  Luo, Lu, et~al.]{chen2023pixart}
Junsong Chen, Jincheng Yu, Chongjian Ge, Lewei Yao, Enze Xie, Yue Wu, Zhongdao
  Wang, James Kwok, Ping Luo, Huchuan Lu, et~al.
\newblock Pixart-$ \alpha $: Fast training of diffusion transformer for
  photorealistic text-to-image synthesis.
\newblock \emph{arXiv preprint arXiv:2310.00426}, 2023{\natexlab{a}}.

\bibitem[Chen et~al.(2025)Chen, Jiang, Wang, Wu, Li, Zhang, Yanai, Chen, and
  Yuan]{chen2025prismlayers}
Junwen Chen, Heyang Jiang, Yanbin Wang, Keming Wu, Ji Li, Chao Zhang, Keiji
  Yanai, Dong Chen, and Yuhui Yuan.
\newblock Prismlayers: Open data for high-quality multi-layer transparent image
  generative models.
\newblock \emph{arXiv preprint arXiv:2505.22523}, 2025.

\bibitem[Chen et~al.(2023{\natexlab{b}})Chen, Zhang, Zeng, Zhang, Zhu, and
  Zhao]{chen2023shikra}
Keqin Chen, Zhao Zhang, Weili Zeng, Richong Zhang, Feng Zhu, and Rui Zhao.
\newblock Shikra: Unleashing multimodal llm's referential dialogue magic.
\newblock \emph{arXiv preprint arXiv:2306.15195}, 2023{\natexlab{b}}.

\bibitem[Chen et~al.(2023{\natexlab{c}})Chen, Li, Dong, Zhang, He, Wang, Zhao,
  and Lin]{chen2023sharegpt4v}
Lin Chen, Jinsong Li, Xiaoyi Dong, Pan Zhang, Conghui He, Jiaqi Wang, Feng
  Zhao, and Dahua Lin.
\newblock Sharegpt4v: Improving large multi-modal models with better captions.
\newblock \emph{arXiv preprint arXiv:2311.12793}, 2023{\natexlab{c}}.

\bibitem[Chen et~al.(2015)Chen, Fang, Lin, Vedantam, Gupta, Doll{\'a}r, and
  Zitnick]{chen2015cococaption}
Xinlei Chen, Hao Fang, Tsung-Yi Lin, Ramakrishna Vedantam, Saurabh Gupta, Piotr
  Doll{\'a}r, and C~Lawrence Zitnick.
\newblock Microsoft coco captions: Data collection and evaluation server.
\newblock \emph{arXiv preprint arXiv:1504.00325}, 2015.

\bibitem[Cherti et~al.(2023)Cherti, Beaumont, Wightman, Wortsman, Ilharco,
  Gordon, Schuhmann, Schmidt, and Jitsev]{cherti2023reproducible}
Mehdi Cherti, Romain Beaumont, Ross Wightman, Mitchell Wortsman, Gabriel
  Ilharco, Cade Gordon, Christoph Schuhmann, Ludwig Schmidt, and Jenia Jitsev.
\newblock Reproducible scaling laws for contrastive language-image learning.
\newblock In \emph{Proceedings of the IEEE/CVF Conference on Computer Vision
  and Pattern Recognition}, pages 2818--2829, 2023.

\bibitem[Dai et~al.(2023)Dai, Li, Li, Tiong, Zhao, Wang, Li, Fung, and
  Hoi]{instructblip}
Wenliang Dai, Junnan Li, Dongxu Li, Anthony Meng~Huat Tiong, Junqi Zhao,
  Weisheng Wang, Boyang Li, Pascale Fung, and Steven Hoi.
\newblock Instructblip: Towards general-purpose vision-language models with
  instruction tuning, 2023.

\bibitem[Deng et~al.(2009)Deng, Dong, Socher, Li, Li, and
  Fei-Fei]{deng2009imagenet}
Jia Deng, Wei Dong, Richard Socher, Li-Jia Li, Kai Li, and Li Fei-Fei.
\newblock Imagenet: A large-scale hierarchical image database.
\newblock In \emph{2009 IEEE conference on computer vision and pattern
  recognition}, pages 248--255. Ieee, 2009.

\bibitem[Dong et~al.(2023)Dong, Bao, Zheng, Zhang, Chen, Yang, Zeng, Zhang,
  Yuan, Chen, et~al.]{dong2023maskclip}
Xiaoyi Dong, Jianmin Bao, Yinglin Zheng, Ting Zhang, Dongdong Chen, Hao Yang,
  Ming Zeng, Weiming Zhang, Lu Yuan, Dong Chen, et~al.
\newblock Maskclip: Masked self-distillation advances contrastive
  language-image pretraining.
\newblock In \emph{Proceedings of the IEEE/CVF Conference on Computer Vision
  and Pattern Recognition}, pages 10995--11005, 2023.

\bibitem[Fan et~al.(2025)Fan, Lu, Dai, Yu, Xiao, Dou, Dong, Ma, and
  Wang]{fan2025go}
Ke Fan, Shunlin Lu, Minyue Dai, Runyi Yu, Lixing Xiao, Zhiyang Dou, Junting
  Dong, Lizhuang Ma, and Jingbo Wang.
\newblock Go to zero: Towards zero-shot motion generation with million-scale
  data.
\newblock \emph{arXiv preprint arXiv:2507.07095}, 2025.

\bibitem[Fan et~al.(2024)Fan, Krishnan, Isola, Katabi, and
  Tian]{fan2024improving}
Lijie Fan, Dilip Krishnan, Phillip Isola, Dina Katabi, and Yonglong Tian.
\newblock Improving clip training with language rewrites.
\newblock \emph{Advances in Neural Information Processing Systems}, 36, 2024.

\bibitem[Garg et~al.(2024)Garg, Burns, Ayan, Bitton, Montgomery, Onoe, Bunner,
  Krishna, Baldridge, and Soricut]{garg2024imageinwords}
Roopal Garg, Andrea Burns, Burcu~Karagol Ayan, Yonatan Bitton, Ceslee
  Montgomery, Yasumasa Onoe, Andrew Bunner, Ranjay Krishna, Jason Baldridge,
  and Radu Soricut.
\newblock Imageinwords: Unlocking hyper-detailed image descriptions.
\newblock \emph{arXiv preprint arXiv:2405.02793}, 2024.

\bibitem[Gu et~al.(2021)Gu, Lin, Kuo, and Cui]{gu2021open}
Xiuye Gu, Tsung-Yi Lin, Weicheng Kuo, and Yin Cui.
\newblock Open-vocabulary object detection via vision and language knowledge
  distillation.
\newblock \emph{arXiv preprint arXiv:2104.13921}, 2021.

\bibitem[Hammoud et~al.(2024)Hammoud, Itani, Pizzati, Torr, Bibi, and
  Ghanem]{hammoud2024synthclip}
Hasan Abed Al~Kader Hammoud, Hani Itani, Fabio Pizzati, Philip Torr, Adel Bibi,
  and Bernard Ghanem.
\newblock Synthclip: Are we ready for a fully synthetic clip training?
\newblock \emph{arXiv preprint arXiv:2402.01832}, 2024.

\bibitem[He et~al.(2020)He, Fan, Wu, Xie, and Girshick]{he2020momentum}
Kaiming He, Haoqi Fan, Yuxin Wu, Saining Xie, and Ross Girshick.
\newblock Momentum contrast for unsupervised visual representation learning.
\newblock In \emph{Proceedings of the IEEE/CVF conference on computer vision
  and pattern recognition}, pages 9729--9738, 2020.

\bibitem[He et~al.(2022)He, Chen, Xie, Li, Doll{\'a}r, and
  Girshick]{he2022masked}
Kaiming He, Xinlei Chen, Saining Xie, Yanghao Li, Piotr Doll{\'a}r, and Ross
  Girshick.
\newblock Masked autoencoders are scalable vision learners.
\newblock In \emph{Proceedings of the IEEE/CVF conference on computer vision
  and pattern recognition}, pages 16000--16009, 2022.

\bibitem[Hendrycks et~al.(2021)Hendrycks, Zhao, Basart, Steinhardt, and
  Song]{hendrycks2021natural}
Dan Hendrycks, Kevin Zhao, Steven Basart, Jacob Steinhardt, and Dawn Song.
\newblock Natural adversarial examples.
\newblock In \emph{Proceedings of the IEEE/CVF conference on computer vision
  and pattern recognition}, pages 15262--15271, 2021.

\bibitem[Hu et~al.(2025)Hu, Wang, Chen, Zhang, Wang, Li, and Nan]{DynamicID}
Xirui Hu, Jiahao Wang, Hao Chen, Weizhan Zhang, Benqi Wang, Yikun Li, and
  Haishun Nan.
\newblock Dynamicid: Zero-shot multi-id image personalization with flexible
  facial editability.
\newblock 2025.

\bibitem[Jain et~al.(2024)Jain, Yang, and Shi]{jain2024vcoder}
Jitesh Jain, Jianwei Yang, and Humphrey Shi.
\newblock Vcoder: Versatile vision encoders for multimodal large language
  models.
\newblock In \emph{Proceedings of the IEEE/CVF Conference on Computer Vision
  and Pattern Recognition}, pages 27992--28002, 2024.

\bibitem[Jia et~al.(2021)Jia, Yang, Xia, Chen, Parekh, Pham, Le, Sung, Li, and
  Duerig]{jia2021scaling}
Chao Jia, Yinfei Yang, Ye Xia, Yi-Ting Chen, Zarana Parekh, Hieu Pham, Quoc Le,
  Yun-Hsuan Sung, Zhen Li, and Tom Duerig.
\newblock Scaling up visual and vision-language representation learning with
  noisy text supervision.
\newblock In \emph{International conference on machine learning}, pages
  4904--4916. PMLR, 2021.

\bibitem[Krishna et~al.(2017)Krishna, Zhu, Groth, Johnson, Hata, Kravitz, Chen,
  Kalantidis, Li, Shamma, et~al.]{krishna2017visual}
Ranjay Krishna, Yuke Zhu, Oliver Groth, Justin Johnson, Kenji Hata, Joshua
  Kravitz, Stephanie Chen, Yannis Kalantidis, Li-Jia Li, David~A Shamma, et~al.
\newblock Visual genome: Connecting language and vision using crowdsourced
  dense image annotations.
\newblock \emph{International journal of computer vision}, 123:\penalty0
  32--73, 2017.

\bibitem[Krizhevsky and Hinton(2009)]{2009Learning}
A. Krizhevsky and G. Hinton.
\newblock Learning multiple layers of features from tiny images.
\newblock \emph{Handbook of Systemic Autoimmune Diseases}, 1\penalty0 (4),
  2009.

\bibitem[Lai et~al.(2025)Lai, Zhang, Zhang, Wu, Bai, Timofeev, Du, Gan, Shan,
  Chuah, et~al.]{lai2025veclip}
Zhengfeng Lai, Haotian Zhang, Bowen Zhang, Wentao Wu, Haoping Bai, Aleksei
  Timofeev, Xianzhi Du, Zhe Gan, Jiulong Shan, Chen-Nee Chuah, et~al.
\newblock Veclip: Improving clip training via visual-enriched captions.
\newblock In \emph{European Conference on Computer Vision}, pages 111--127.
  Springer, 2025.

\bibitem[Lan et~al.(2024)Lan, Chen, Ke, Wang, Feng, and
  Zhang]{lan2024proxyclip}
Mengcheng Lan, Chaofeng Chen, Yiping Ke, Xinjiang Wang, Litong Feng, and Wayne
  Zhang.
\newblock Proxyclip: Proxy attention improves clip for open-vocabulary
  segmentation.
\newblock \emph{arXiv preprint arXiv:2408.04883}, 2024.

\bibitem[Li et~al.(2022{\natexlab{a}})Li, Weinberger, Belongie, Koltun, and
  Ranftl]{li2022language}
Boyi Li, Kilian~Q Weinberger, Serge Belongie, Vladlen Koltun, and Ren{\'e}
  Ranftl.
\newblock Language-driven semantic segmentation.
\newblock \emph{arXiv preprint arXiv:2201.03546}, 2022{\natexlab{a}}.

\bibitem[Li et~al.(2022{\natexlab{b}})Li, Li, Xiong, and Hoi]{li2022blip}
Junnan Li, Dongxu Li, Caiming Xiong, and Steven Hoi.
\newblock Blip: Bootstrapping language-image pre-training for unified
  vision-language understanding and generation.
\newblock In \emph{International conference on machine learning}, pages
  12888--12900. PMLR, 2022{\natexlab{b}}.

\bibitem[Li et~al.(2023{\natexlab{a}})Li, Li, Savarese, and Hoi]{li2023blip}
Junnan Li, Dongxu Li, Silvio Savarese, and Steven Hoi.
\newblock Blip-2: Bootstrapping language-image pre-training with frozen image
  encoders and large language models.
\newblock In \emph{International conference on machine learning}, pages
  19730--19742. PMLR, 2023{\natexlab{a}}.

\bibitem[Li et~al.(2022{\natexlab{c}})Li, Zhang, Zhang, Yang, Li, Zhong, Wang,
  Yuan, Zhang, Hwang, et~al.]{li2022grounded}
Liunian~Harold Li, Pengchuan Zhang, Haotian Zhang, Jianwei Yang, Chunyuan Li,
  Yiwu Zhong, Lijuan Wang, Lu Yuan, Lei Zhang, Jenq-Neng Hwang, et~al.
\newblock Grounded language-image pre-training.
\newblock In \emph{Proceedings of the IEEE/CVF Conference on Computer Vision
  and Pattern Recognition}, pages 10965--10975, 2022{\natexlab{c}}.

\bibitem[Li et~al.(2023{\natexlab{b}})Li, Fan, Hu, Feichtenhofer, and
  He]{li2023scaling}
Yanghao Li, Haoqi Fan, Ronghang Hu, Christoph Feichtenhofer, and Kaiming He.
\newblock Scaling language-image pre-training via masking.
\newblock In \emph{Proceedings of the IEEE/CVF Conference on Computer Vision
  and Pattern Recognition}, pages 23390--23400, 2023{\natexlab{b}}.

\bibitem[Li et~al.(2024)Li, Li, Zeng, Hou, and Cheng]{li2024cascade}
Yunheng Li, ZhongYu Li, Quansheng Zeng, Qibin Hou, and Ming-Ming Cheng.
\newblock Cascade-clip: Cascaded vision-language embeddings alignment for
  zero-shot semantic segmentation.
\newblock \emph{arXiv preprint arXiv:2406.00670}, 2024.

\bibitem[Liu et~al.(2024{\natexlab{a}})Liu, Li, Li, and Lee]{liu2024improved}
Haotian Liu, Chunyuan Li, Yuheng Li, and Yong~Jae Lee.
\newblock Improved baselines with visual instruction tuning.
\newblock In \emph{Proceedings of the IEEE/CVF Conference on Computer Vision
  and Pattern Recognition}, pages 26296--26306, 2024{\natexlab{a}}.

\bibitem[Liu et~al.(2024{\natexlab{b}})Liu, Li, Li, Li, Zhang, Shen, and
  Lee]{liu2024llavanext}
Haotian Liu, Chunyuan Li, Yuheng Li, Bo Li, Yuanhan Zhang, Sheng Shen, and
  Yong~Jae Lee.
\newblock Llava-next: Improved reasoning, ocr, and world knowledge,
  2024{\natexlab{b}}.

\bibitem[Loshchilov and Hutter(2017)]{loshchilov2017adamw}
Ilya Loshchilov and Frank Hutter.
\newblock Decoupled weight decay regularization.
\newblock \emph{arXiv preprint arXiv:1711.05101}, 2017.

\bibitem[Mukhoti et~al.(2023)Mukhoti, Lin, Poursaeed, Wang, Shah, Torr, and
  Lim]{mukhoti2023open}
Jishnu Mukhoti, Tsung-Yu Lin, Omid Poursaeed, Rui Wang, Ashish Shah, Philip~HS
  Torr, and Ser-Nam Lim.
\newblock Open vocabulary semantic segmentation with patch aligned contrastive
  learning.
\newblock In \emph{Proceedings of the IEEE/CVF Conference on Computer Vision
  and Pattern Recognition}, pages 19413--19423, 2023.

\bibitem[Najdenkoska et~al.(2024)Najdenkoska, Derakhshani, Asano, van Noord,
  Worring, and Snoek]{najdenkoska2024tulip}
Ivona Najdenkoska, Mohammad~Mahdi Derakhshani, Yuki~M Asano, Nanne van Noord,
  Marcel Worring, and Cees~GM Snoek.
\newblock Tulip: Token-length upgraded clip.
\newblock \emph{arXiv preprint arXiv:2410.10034}, 2024.

\bibitem[Ni et~al.(2025)Ni, Wang, Zhu, Wang, Li, Zhao, Li, Qin, Huang, and
  Mei]{ni2025wonderturbo}
Chaojun Ni, Xiaofeng Wang, Zheng Zhu, Weijie Wang, Haoyun Li, Guosheng Zhao,
  Jie Li, Wenkang Qin, Guan Huang, and Wenjun Mei.
\newblock Wonderturbo: Generating interactive 3d world in 0.72 seconds, 2025.

\bibitem[Ordonez et~al.(2011)Ordonez, Kulkarni, and Berg]{ordonez2011im2text}
Vicente Ordonez, Girish Kulkarni, and Tamara Berg.
\newblock Im2text: Describing images using 1 million captioned photographs.
\newblock \emph{Advances in neural information processing systems}, 24, 2011.

\bibitem[Peebles and Xie(2023)]{peebles2023scalable}
William Peebles and Saining Xie.
\newblock Scalable diffusion models with transformers.
\newblock In \emph{Proceedings of the IEEE/CVF International Conference on
  Computer Vision}, pages 4195--4205, 2023.

\bibitem[Peng et~al.(2025)Peng, Xiao, Wu, Liao, Chen, Lin, Huang, Li, and
  Yuan]{peng2025bizgen}
Yuyang Peng, Shishi Xiao, Keming Wu, Qisheng Liao, Bohan Chen, Kevin Lin,
  Danqing Huang, Ji Li, and Yuhui Yuan.
\newblock Bizgen: Advancing article-level visual text rendering for
  infographics generation.
\newblock In \emph{Proceedings of the Computer Vision and Pattern Recognition
  Conference}, pages 23615--23624, 2025.

\bibitem[Plummer et~al.(2015)Plummer, Wang, Cervantes, Caicedo, Hockenmaier,
  and Lazebnik]{plummer2015flickr30k}
Bryan~A Plummer, Liwei Wang, Chris~M Cervantes, Juan~C Caicedo, Julia
  Hockenmaier, and Svetlana Lazebnik.
\newblock Flickr30k entities: Collecting region-to-phrase correspondences for
  richer image-to-sentence models.
\newblock In \emph{Proceedings of the IEEE international conference on computer
  vision}, pages 2641--2649, 2015.

\bibitem[Radford et~al.(2021)Radford, Kim, Hallacy, Ramesh, Goh, Agarwal,
  Sastry, Askell, Mishkin, Clark, et~al.]{radford2021learning}
Alec Radford, Jong~Wook Kim, Chris Hallacy, Aditya Ramesh, Gabriel Goh,
  Sandhini Agarwal, Girish Sastry, Amanda Askell, Pamela Mishkin, Jack Clark,
  et~al.
\newblock Learning transferable visual models from natural language
  supervision.
\newblock In \emph{International conference on machine learning}, pages
  8748--8763. PMLR, 2021.

\bibitem[Recht et~al.(2019)Recht, Roelofs, Schmidt, and
  Shankar]{recht2019imagenet}
Benjamin Recht, Rebecca Roelofs, Ludwig Schmidt, and Vaishaal Shankar.
\newblock Do imagenet classifiers generalize to imagenet?
\newblock In \emph{International conference on machine learning}, pages
  5389--5400. PMLR, 2019.

\bibitem[Rombach et~al.(2022)Rombach, Blattmann, Lorenz, Esser, and
  Ommer]{Rombach_Blattmann_Lorenz_Esser_Ommer_2022}
Robin Rombach, Andreas Blattmann, Dominik Lorenz, Patrick Esser, and Bjorn
  Ommer.
\newblock High-resolution image synthesis with latent diffusion models.
\newblock In \emph{2022 IEEE/CVF Conference on Computer Vision and Pattern
  Recognition (CVPR)}, 2022.

\bibitem[Sharma et~al.(2018)Sharma, Ding, Goodman, and
  Soricut]{sharma2018conceptual}
Piyush Sharma, Nan Ding, Sebastian Goodman, and Radu Soricut.
\newblock Conceptual captions: A cleaned, hypernymed, image alt-text dataset
  for automatic image captioning.
\newblock In \emph{Proceedings of the 56th Annual Meeting of the Association
  for Computational Linguistics (Volume 1: Long Papers)}, pages 2556--2565,
  2018.

\bibitem[Shi et~al.(2025)Shi, Zhao, Wang, Zhang, Wang, Li, Dai, Zou, Xiong,
  Tian, et~al.]{shi2025umg}
Bowen Shi, Peisen Zhao, Zichen Wang, Yuhang Zhang, Yaoming Wang, Jin Li, Wenrui
  Dai, Junni Zou, Hongkai Xiong, Qi Tian, et~al.
\newblock Umg-clip: A unified multi-granularity vision generalist for
  open-world understanding.
\newblock In \emph{European Conference on Computer Vision}, pages 259--277.
  Springer, 2025.

\bibitem[Sim{\'e}oni et~al.(2021)Sim{\'e}oni, Puy, Vo, Roburin, Gidaris,
  Bursuc, P{\'e}rez, Marlet, and Ponce]{simeoni2021localizing}
Oriane Sim{\'e}oni, Gilles Puy, Huy~V Vo, Simon Roburin, Spyros Gidaris, Andrei
  Bursuc, Patrick P{\'e}rez, Renaud Marlet, and Jean Ponce.
\newblock Localizing objects with self-supervised transformers and no labels.
\newblock \emph{arXiv preprint arXiv:2109.14279}, 2021.

\bibitem[Sun et~al.(2023)Sun, Fang, Wu, Wang, and Cao]{sun2023eva}
Quan Sun, Yuxin Fang, Ledell Wu, Xinlong Wang, and Yue Cao.
\newblock Eva-clip: Improved training techniques for clip at scale.
\newblock \emph{arXiv preprint arXiv:2303.15389}, 2023.

\bibitem[Thomee et~al.(2016)Thomee, Shamma, Friedland, Elizalde, Ni, Poland,
  Borth, and Li]{thomee2016yfcc100m}
Bart Thomee, David~A Shamma, Gerald Friedland, Benjamin Elizalde, Karl Ni,
  Douglas Poland, Damian Borth, and Li-Jia Li.
\newblock Yfcc100m: The new data in multimedia research.
\newblock \emph{Communications of the ACM}, 59\penalty0 (2):\penalty0 64--73,
  2016.

\bibitem[Urbanek et~al.(2024)Urbanek, Bordes, Astolfi, Williamson, Sharma, and
  Romero-Soriano]{urbanek2024picture}
Jack Urbanek, Florian Bordes, Pietro Astolfi, Mary Williamson, Vasu Sharma, and
  Adriana Romero-Soriano.
\newblock A picture is worth more than 77 text tokens: Evaluating clip-style
  models on dense captions.
\newblock In \emph{Proceedings of the IEEE/CVF Conference on Computer Vision
  and Pattern Recognition}, pages 26700--26709, 2024.

\bibitem[Wang et~al.(2023{\natexlab{a}})Wang, Zhou, Shou, and
  Yan]{wang2023position}
Jinpeng Wang, Pan Zhou, Mike~Zheng Shou, and Shuicheng Yan.
\newblock Position-guided text prompt for vision-language pre-training.
\newblock In \emph{Proceedings of the IEEE/CVF Conference on Computer Vision
  and Pattern Recognition}, pages 23242--23251, 2023{\natexlab{a}}.

\bibitem[Wang et~al.(2023{\natexlab{b}})Wang, Bao, Dong, Bjorck, Peng, Liu,
  Aggarwal, Mohammed, Singhal, Som, et~al.]{wang2023image}
Wenhui Wang, Hangbo Bao, Li Dong, Johan Bjorck, Zhiliang Peng, Qiang Liu, Kriti
  Aggarwal, Owais~Khan Mohammed, Saksham Singhal, Subhojit Som, et~al.
\newblock Image as a foreign language: Beit pretraining for vision and
  vision-language tasks.
\newblock In \emph{Proceedings of the IEEE/CVF Conference on Computer Vision
  and Pattern Recognition}, pages 19175--19186, 2023{\natexlab{b}}.

\bibitem[Wu et~al.(2024)Wu, Zheng, Ma, Lu, Guo, Zhang, Chen, Guo, Shen, and
  Zha]{wu2024lotlip}
Wei Wu, Kecheng Zheng, Shuailei Ma, Fan Lu, Yuxin Guo, Yifei Zhang, Wei Chen,
  Qingpei Guo, Yujun Shen, and Zheng-Jun Zha.
\newblock Lotlip: Improving language-image pre-training for long text
  understanding.
\newblock \emph{arXiv preprint arXiv:2410.05249}, 2024.

\bibitem[Xiao et~al.(2024)Xiao, Kim, Georgescu, Akata, and
  Alaniz]{xiao2024flair}
Rui Xiao, Sanghwan Kim, Mariana-Iuliana Georgescu, Zeynep Akata, and Stephan
  Alaniz.
\newblock Flair: Vlm with fine-grained language-informed image representations.
\newblock \emph{arXiv preprint arXiv:2412.03561}, 2024.

\bibitem[Xu et~al.(2023)Xu, Xie, Tan, Huang, Howes, Sharma, Li, Ghosh,
  Zettlemoyer, and Feichtenhofer]{xu2023demystifying}
Hu Xu, Saining Xie, Xiaoqing~Ellen Tan, Po-Yao Huang, Russell Howes, Vasu
  Sharma, Shang-Wen Li, Gargi Ghosh, Luke Zettlemoyer, and Christoph
  Feichtenhofer.
\newblock Demystifying clip data.
\newblock \emph{arXiv preprint arXiv:2309.16671}, 2023.

\bibitem[Yao et~al.(2021)Yao, Huang, Hou, Lu, Niu, Xu, Liang, Li, Jiang, and
  Xu]{yao2021filip}
Lewei Yao, Runhui Huang, Lu Hou, Guansong Lu, Minzhe Niu, Hang Xu, Xiaodan
  Liang, Zhenguo Li, Xin Jiang, and Chunjing Xu.
\newblock Filip: Fine-grained interactive language-image pre-training.
\newblock \emph{arXiv preprint arXiv:2111.07783}, 2021.

\bibitem[You et~al.(2025{\natexlab{a}})You, Yang, Zhang, Jiang, Yang, and
  Navab]{you2025fbdifffourierbasisguideddiffusion}
Xin You, Runze Yang, Chuyan Zhang, Zhongliang Jiang, Jie Yang, and Nassir
  Navab.
\newblock Fb-diff: Fourier basis-guided diffusion for temporal interpolation of
  4d medical imaging, 2025{\natexlab{a}}.

\bibitem[You et~al.(2025{\natexlab{b}})You, Zhang, Zhang, Yang, and
  Navab]{you2025temporaldifferentialfields4d}
Xin You, Minghui Zhang, Hanxiao Zhang, Jie Yang, and Nassir Navab.
\newblock Temporal differential fields for 4d motion modeling via
  image-to-video synthesis, 2025{\natexlab{b}}.

\bibitem[Yu et~al.(2024)Yu, Sun, Zhang, Cui, Zhang, Cao, Wang, and
  Liu]{yu2024capsfusion}
Qiying Yu, Quan Sun, Xiaosong Zhang, Yufeng Cui, Fan Zhang, Yue Cao, Xinlong
  Wang, and Jingjing Liu.
\newblock Capsfusion: Rethinking image-text data at scale.
\newblock In \emph{Proceedings of the IEEE/CVF Conference on Computer Vision
  and Pattern Recognition}, pages 14022--14032, 2024.

\bibitem[Zhai et~al.(2023)Zhai, Mustafa, Kolesnikov, and
  Beyer]{zhai2023sigmoid}
Xiaohua Zhai, Basil Mustafa, Alexander Kolesnikov, and Lucas Beyer.
\newblock Sigmoid loss for language image pre-training.
\newblock In \emph{Proceedings of the IEEE/CVF international conference on
  computer vision}, pages 11975--11986, 2023.

\bibitem[Zhang et~al.(2024{\natexlab{a}})Zhang, Zhang, Dong, Zang, and
  Wang]{zhang2024long}
Beichen Zhang, Pan Zhang, Xiaoyi Dong, Yuhang Zang, and Jiaqi Wang.
\newblock Long-clip: Unlocking the long-text capability of clip.
\newblock \emph{arXiv preprint arXiv:2403.15378}, 2024{\natexlab{a}}.

\bibitem[Zhang et~al.(2024{\natexlab{b}})Zhang, Huang, Jin, and
  Lu]{zhang2024vision}
Jingyi Zhang, Jiaxing Huang, Sheng Jin, and Shijian Lu.
\newblock Vision-language models for vision tasks: A survey.
\newblock \emph{IEEE Transactions on Pattern Analysis and Machine
  Intelligence}, 2024{\natexlab{b}}.

\bibitem[Zhang et~al.(2024{\natexlab{c}})Zhang, Guo, Wang, and
  Hu]{zhang2024exploring}
Yi Zhang, Meng-Hao Guo, Miao Wang, and Shi-Min Hu.
\newblock Exploring regional clues in clip for zero-shot semantic segmentation.
\newblock In \emph{Proceedings of the IEEE/CVF Conference on Computer Vision
  and Pattern Recognition}, pages 3270--3280, 2024{\natexlab{c}}.

\bibitem[Zheng et~al.(2025)Zheng, Zhang, Wu, Lu, Ma, Jin, Chen, and
  Shen]{zheng2025dreamlip}
Kecheng Zheng, Yifei Zhang, Wei Wu, Fan Lu, Shuailei Ma, Xin Jin, Wei Chen, and
  Yujun Shen.
\newblock Dreamlip: Language-image pre-training with long captions.
\newblock In \emph{European Conference on Computer Vision}, pages 73--90.
  Springer, 2025.

\bibitem[Zhou et~al.(2023)Zhou, Lei, Zhang, Liu, and Liu]{zhou2023zegclip}
Ziqin Zhou, Yinjie Lei, Bowen Zhang, Lingqiao Liu, and Yifan Liu.
\newblock Zegclip: Towards adapting clip for zero-shot semantic segmentation.
\newblock In \emph{Proceedings of the IEEE/CVF Conference on Computer Vision
  and Pattern Recognition}, pages 11175--11185, 2023.

\end{thebibliography}
}
\clearpage
\setcounter{page}{1}
\maketitlesupplementary

\section{Prompting Templates for Long-text Caption Synthesis}
To ensure the diversity of the synthesis long-text captions, we have set up multiple prompts to instruct Llama3-LLaVA-NeXT-8b~\cite{liu2024llavanext} to generate long-text captions with detailed descriptions. During the re-caption process, samples are randomly taken from the following 20 prompts.\\
\label{prompts}
\newcommand{\prompt}[1]{\textcolor{black}{\small{\texttt{#1}}}}
\noindent\prompt{1. Provide a comprehensive description of this image, including all visual elements, their spatial relationships, and the overall atmosphere.} \\
\prompt{2. Generate a detailed caption explaining what's happening in this image, covering actions, subjects, environment, and temporal context.} \\
\prompt{3. Analyze this image in detail, describing the main subjects, background, lighting, colors, and composition.} \\
\prompt{4. Write an extensive caption that captures both the explicit visual content and implicit context or story behind this image.} \\
\prompt{5. Describe this image as if explaining it to someone who cannot see it, including all relevant details and visual nuances.} \\
\prompt{6. Break down the scene components in this image, detailing the foreground, middle ground, and background elements.} \\
\prompt{7. Describe the environmental context, lighting conditions, time of day, and weather elements visible in this image.} \\
\prompt{8. Analyze the spatial arrangement and relationships between all objects and subjects in this image.} \\
\prompt{9. Detail the setting of this scene, including architectural elements, natural features, and atmospheric conditions.} \\
\prompt{10. Explain the visual dynamics of this scene, including movement, direction, and flow of elements.} \\
\prompt{11. Elaborate on the image's details such as the objects' textures, the direction of shadows, and how they contribute to the overall look.} \\
\prompt{12. Describe the image from top to bottom and left to right, highlighting every element and its significance within the frame.} \\
\prompt{13. Generate a caption that delves into the emotional undertones suggested by the image's colors, expressions of the subjects, and the setting.} \\
\prompt{14. Analyze the image to explain how the placement of elements affects the flow and balance within the visual space.} \\
\prompt{15. Write a detailed description of the image that includes the sizes of the objects relative to each other and their proximity.} \\
\prompt{16. Describe the image in terms of the contrast between light and dark areas and how it shapes the perception of the scene.} \\
\prompt{17. Generate a caption that interprets the possible narrative connections between different elements in the image.} \\
\prompt{18. Analyze the image to explain how the colors interact with each other and what mood they create together.} \\
\prompt{19. Write a detailed description of the image that covers the small details often overlooked, like tiny patterns on objects.} \\
\prompt{20. Describe the image by focusing on the perspective used and how it makes the viewer experience the scene.} \\

\section{Abnormal Synthesized Captions}
\label{sec:incorrect caption examples}
\begin{figure}[t] 
	\centering  
	\includegraphics[width=1.0\linewidth]{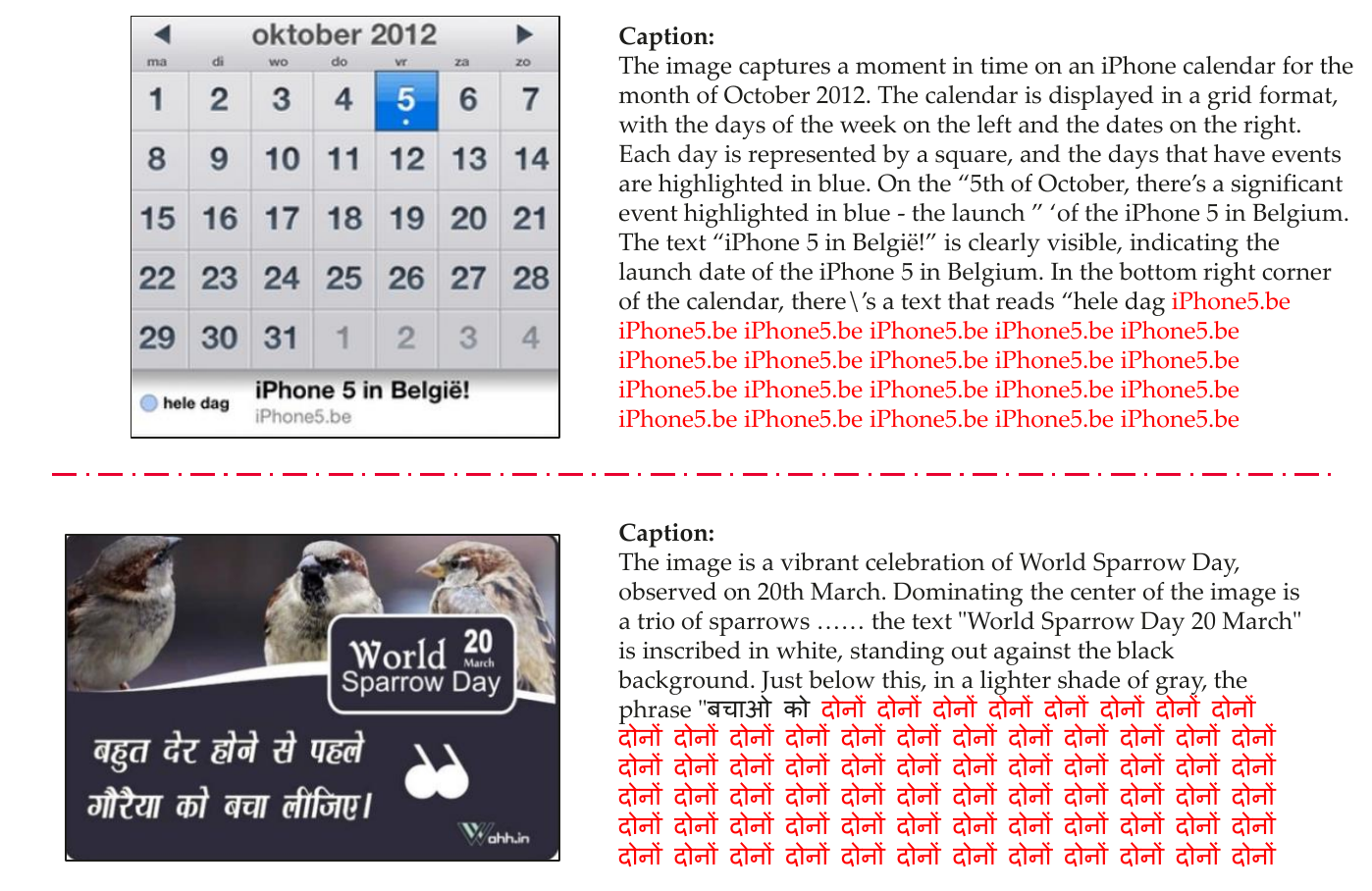}
	\caption{Some incorrect examples from our re-captioned dataset. Both images are wrong captioned with repeating words.}
	\label{fig:incorrect dataset}
\end{figure}

While synthesized captions provide detailed descriptions, MLLMs usually bring hallucination elements. We apply a simple filtering method on captions to reduce repeated words, meaningless sentences, and short results. \cref{fig:incorrect dataset} shows some abnormal synthesized captions that have been cleaned out from our training datasets.\
\section{Details of the Setup}
\label{sec:train_hyper}

\subsection{Details of the training datasets}
Our model's training corpus comprises six distinct datasets, as enumerated in \cref{tab:training dataset detail}. The ShareGPT4V~\cite{chen2023sharegpt4v} dataset, previously employed in Long-CLIP~\cite{zhang2024long} implementation, exhibits exceptional annotation quality. The remaining five established datasets, including CC3M~\cite{sharma2018conceptual}, VisualGenome~\cite{krishna2017visual}, SBU~\cite{ordonez2011im2text}, CC12M~\cite{sharma2018conceptual}, and YFCC15M~\cite{thomee2016yfcc100m}, underwent our custom annotation process, utilizing the previously described Llama3-LLaVA-NeXT-8b~\cite{liu2024llavanext} model for generating extensive long-text caption synthesis. These datasets were systematically organized into three distinct scales: 5M, 15M, and 30M for training purposes. 

\cref{tab:training dataset detail} provides comprehensive statistics, including the quantity of image-text pairs, sentences per text, and tokens per text. Comparative analysis reveals that our annotated datasets demonstrate marginally lower text lengths relative to ShareGPT4V~\cite{chen2023sharegpt4v}, a characteristic potentially attributed to model-specific limitations, which may impose certain constraints on our model's performance upper bound.

\begin{table}[t]
    \centering
    \large
    \resizebox{\linewidth}{!}{
    \begin{tabular}{c| c| c| c}
    \toprule
    Dataset&Image-text pairs&Sentences per Text &Tokens per Text  \\
    \midrule
        CC3M~\cite{sharma2018conceptual}& 2760314 & 6.31 & 116.96 \\
        VisualGenome~\cite{krishna2017visual}& 107653 & 6.52 & 117.68 \\
        ShareGPT4V~\cite{chen2023sharegpt4v}& 1246901 & 9.22 & 172.94 \\
        SBU~\cite{ordonez2011im2text} & 835333 & 6.01 & 110.33 \\
        CC12M~\cite{sharma2018conceptual}& 8523767 & 6.84 & 131.13 \\
        YFCC15M~\cite{thomee2016yfcc100m} & 14994664 & 6.14 & 115.38 \\
    \bottomrule
    \end{tabular}
    }
    \caption{Details of training datasets. We cleaned the data, so the number of image-text pairs is slightly less than that of the original datasets.}
    \label{tab:training dataset detail}
\end{table}

\begin{figure}[t] 
	\centering  
	\includegraphics[width=1.0\linewidth]{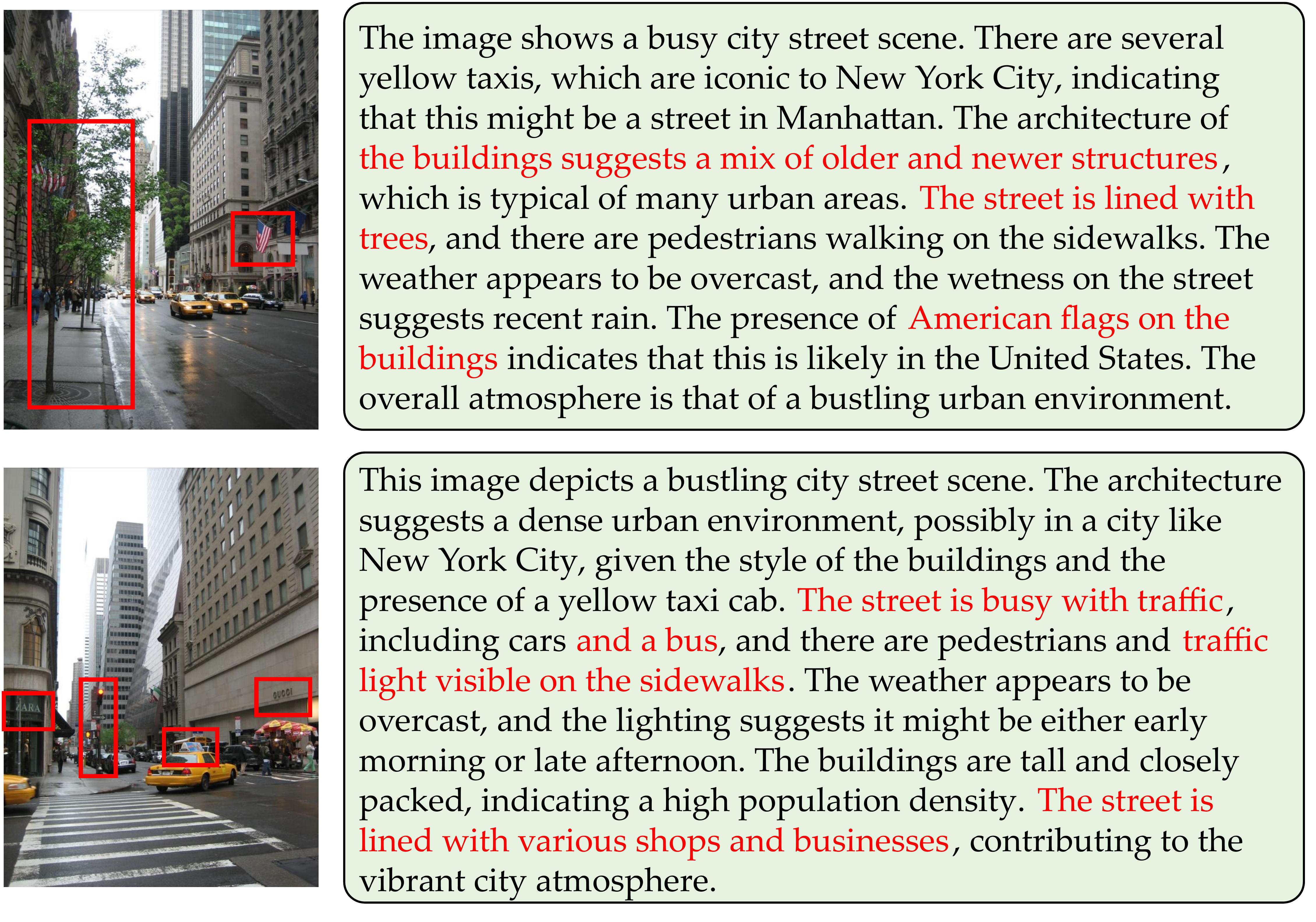}
	\caption{Some examples from our re-captioned dataset. The captions of two similar images are both synthesized by Llama3-LLaVA-NeXT-8b. The key attributes to distinguish these images are marked in red, and highlighted by the red boxes in the images.}
	\label{fig:dataset}
\end{figure}

We randomly selected two visually similar images from the VisualGenome~\cite{krishna2017visual} dataset, with their corresponding synthesized long-text captions presented in \cref{fig:dataset}. Despite strong similarities in architectural style, scene elements, weather conditions, and lighting characteristics between these images, our synthesized captions demonstrate precise differentiation of fine-grained details. The text segments highlighted in red accurately delineate the fine-grained visual information contained within the red-bounded regions of the respective images.

\subsection{Details of the retrieval tasks}
To evaluate our model's cross-modal retrieval capabilities, we conducted experiments on both long-text and short-text retrieval tasks. Traditional retrieval evaluations, primarily conducted onCOCO~\cite{chen2015cococaption} and Flickr30k~\cite{plummer2015flickr30k} with an average text length below 15 tokens, are predominantly focused on short-text image-text retrieval capabilities.

For comprehensive long-text retrieval assessment, we adopted the experimental configurations from established works: Urban-1k~\cite{zhang2024long} and ShareGPT4V~\cite{chen2023sharegpt4v} settings from Long-CLIP~\cite{zhang2024long}, and DCI~\cite{urbanek2024picture} and IIW~\cite{garg2024imageinwords} configurations from LoTLIP~\cite{wu2024lotlip}, ensuring fair comparative analysis. \cref{tab:retrieval dataset detail} presents detailed statistical characteristics of these benchmark datasets.
\begin{table}[t]
    \centering 
    \large
    \resizebox{\linewidth}{!}{
    \begin{tabular}{c| c| c| c| c|c}
    \toprule
    &Dataset&Images&texts&Sentences per Text &Tokens per Text  \\
    \midrule
    \multirow{4}{*}{Long-Text} & ShareGPT4V~\cite{chen2023sharegpt4v} & 1000 & 1000& 8.15 & 173.24\\
        &Urban-1k~\cite{zhang2024long}&1000&1000& 7.088 & 129.24 \\
        &DCI~\cite{urbanek2024picture} & 7805&7805&10.81&172.73\\
        &IIW~\cite{garg2024imageinwords}&612&612&10.16&39.73\\
    \midrule
    \multirow{2}{*}{Short-Text} &COCO~\cite{chen2015cococaption} &5000&25000&1.0&11.77\\
        &Flickr30k~\cite{plummer2015flickr30k}&1000&5000&1.0&14.03\\
    \bottomrule
    \end{tabular}
    }
    \caption{Dataset details of retrieval tasks.}
    \label{tab:retrieval dataset detail}
\end{table}

\subsection{Hyperparameters}
Training hyperparameters of \ours are presented in \cref{tab:train_hyper}. For a fair comparison, our training hyperparameters are consistent with Long-CLIP~\cite{zhang2024long}.

 \begin{table*}[t]
    \centering 
    \resizebox{0.88\linewidth}{!}{
    \begin{tabular}{c| c| c| c| c|c| c| c| c| c|c}
    \toprule
    \multirow{2}{*}{} &\multirow{2}{*}{Model}
    &\multicolumn{2}{c|}{DCI}  &\multicolumn{2}{c|}{IIW} & \multicolumn{2}{c|}{ShareGPT4V-1k} & \multicolumn{2}{c|}{Urban-1k} &\multirow{2}{*}{Avg.}\\
    && ~I-to-T~ &  ~T-to-I~ &  ~I-to-T~ &  ~T-to-I~ & ~I-to-T~ &  ~T-to-I~ &  ~I-to-T~ &  ~T-to-I~ \\
    \midrule
    \multirow{2}{*}{B/16} & Raw Short Caption &66.2&67.1&\textbf{97.1}&96.7&97.8&97.6&87.7&90.1&87.5\\
        & Synthesis Short Caption& \textbf{67.1} & \textbf{67.5} & 96.9 & \textbf{96.7}&\textbf{98.1} & \textbf{97.9} & \textbf{88.0} & \textbf{90.8}& \textbf{87.9}\\

    \midrule
    \multirow{2}{*}{L/14} & Raw Short Caption & 66.5&69.1&\textbf{97.3}&97.0&97.4&97.6&87.9&92.6&88.1\\
    & Synthesis Short Caption&\textbf{68.1} & \textbf{69.9} & 97.1 & \textbf{97.2}&\textbf{98.5} & \textbf{98.0} & \textbf{88.1} & \textbf{93.0}& \textbf{88.7} \\
    \bottomrule
    \end{tabular}
    }
    \caption{Train on 5M synthesis long captions as the long-text input, we compare the performance between the raw short captions and the synthesis short captions as the short-text input. The R@1 of long-text-image retrieval on DCI~\cite{urbanek2024picture}, IIW~\cite{garg2024imageinwords}, ShareGPT4V-1k~\cite{chen2023sharegpt4v}, and Urban-1k~\cite{zhang2024long} datasets. The best results are in \textbf{bold}.}
    \vspace{-1em}
    \label{tab:raw-caption}
\end{table*}

 \begin{table*}[t]
    \centering 
    \resizebox{\textwidth}{!}{
    \begin{tabular}{c| c|  c  c  c|  c c c| c c c |c c c| c}
    \toprule
    \multirow{3}{*}{} &\multirow{3}{*}{Model}
     & \multicolumn{6}{c|}{COCO} & \multicolumn{6}{c|}{Flickr30k} & \multirow{2}{*}{Avg.} \\
    && \multicolumn{3}{c|}{Image-to-Text} &  \multicolumn{3}{c|}{Text-to-Image}&  \multicolumn{3}{c|}{Image-to-Text}&  \multicolumn{3}{c|}{Text-to-Image}&\\
    && R@1~ & R@5~ &R@10~ & R@1~ & R@5~ &R@10~ & R@1~ & R@5~ &R@10~ & R@1~ & R@5~ &R@10~ & R@1\\
    \midrule
    \multirow{2}{*}{B/16} & Raw Short Caption &61.0 &84.5 &90.8 &44.6 &70.4&79.5& 89.2 & 98.4 & 99.7 & 77.4& 94.6 & 97.2 & 68.0\\
        & Synthesis Short Caption &\textbf{61.3} & \textbf{84.9}& \textbf{91.2}& \textbf{47.0}& \textbf{72.4}& \textbf{81.4}& \textbf{89.9} & \textbf{98.8}& \textbf{99.7} & \textbf{78.4} & \textbf{95.2} & \textbf{97.7} &\textbf{69.2}\\ 
    \midrule
    \multirow{2}{*}{L/14} & Raw Short Caption& 62.5 & 85.6 & 91.4 & 48.5 & 73.6 & 82.1 & 92.3 & \textbf{99.3} & 99.7 & 81.7 & 95.9 & 97.9 & 71.2\\
        & Synthesis Short Caption &\textbf{63.2}&\textbf{85.8}&\textbf{91.5}&\textbf{50.5}&\textbf{75.4}&\textbf{83.6}&\textbf{92.5}&99.1&\textbf{99.9}&\textbf{82.5}&\textbf{96.6}&\textbf{98.2}&\textbf{72.1}\\ 
    \bottomrule
    \end{tabular}
    }
    \caption{Train on 5M synthesis long captions as the long-text input, we compare the performance between the raw short captions and the synthesis short captions as the short-text input. Results of short-caption text-image retrieval on the 5k COCO2017~\cite{chen2015cococaption} validation set and the 1k Flickr30K~\cite{plummer2015flickr30k} test set. The best results are in \textbf{bold}.}
    \vspace{-1em}
    \label{tab:raw-caption-2}
\end{table*}

\begin{table}[t]
    \centering
    \tabcolsep=5pt
    \resizebox{0.8\linewidth}{!}{
    \begin{tabular}{lc}
    \toprule
    Configuration & \ours Training \\
    \midrule
    Batch size & 2048 \\
    Training Epoch & 6 \\
    Learning Rate & 1e-6 \\
    Warm-up Steps & 200 \\
    LR Scheduler & cosine \\
    Optimizer & AdamW~\citep{loshchilov2017adamw} \\
    Optimizer hyper-parameters & $\beta_1$, $\beta_2$, $\epsilon$ = 0.9, 0.999, 1e-8 \\
    Weight decay & 1e-2 \\\bottomrule
    \end{tabular}
    }
    \caption{Summary of \ours training hyperparameters.}
    \label{tab:train_hyper}
\end{table}

\begin{table}[h]
    \vspace{-1em}
    \tiny
    \renewcommand{\arraystretch}{0.8}
    \centering 
    \resizebox{\linewidth}{!}{
    \begin{tabular}{c| c| c| c|c |c |c}
    \toprule
    & \multicolumn{2}{c|}{COCO} & \multicolumn{2}{c|}{Urban1k} & \multicolumn{2}{c}{DCI} \\
    & I2T & T2I & I2T & T2I & I2T & T2I \\
    \midrule
    Default & \textbf{62.0} & \textbf{46.7} & \textbf{87.0} & \textbf{86.8} & \textbf{65.1} & \textbf{66.7} \\
    Shared Prompts & 60.7 & 46.0 & 85.2 & 86.1 & 62.9 & 65.5 \\
    R2P & 60.3 & 46.0 & 85.8 & 85.7 & 63.1 & 65.3 \\
    P2R & 61.5 & 46.3 & 86.6 & 86.1 & 64.3 & 66.1 \\
    \midrule
    Short PE (len=77) & 61.2 & 46.3 & 77.8 & 75.4 & 56.3 & 59.1 \\
    Long PE (len=248)& \textbf{62.0} & \textbf{46.7} & \textbf{87.0} & \textbf{86.8} & \textbf{65.1} & \textbf{66.7} \\
    \bottomrule
    \end{tabular}
    }

    \caption{Above: ablations on the efficacy of region prompts and masks. ``Shared Prompts" refers to all the layers utilizing the same shared prompts, ``R2P'' and ``P2R'' denote regional prompts attending to all patch embeddings and vice versa. Bottom: the performance comparison of different position embeddings.}
    \label{tab:suppl_ablation}
\end{table}

\section{Raw Short Caption versus Synthesis Short Caption}
We identified quality limitations in the raw short captions within our training dataset through empirical observation. To address this constraint, we proposed an alternative approach utilizing synthetically generated short captions as model inputs. We conducted comprehensive comparative analyses between models trained on synthetic short captions versus those trained on raw short captions, with results presented in \cref{tab:raw-caption} and \cref{tab:raw-caption-2}. The synthetic short captions were generated by Shikra~\cite{chen2023shikra}. Quantitative evaluations demonstrate that incorporating synthetic short captions into the training dataset yields substantial performance gains, suggesting the effectiveness of our proposed approach.

\section{Visualization of the Effects of Unidirectional Masking and Region Prompts}
\begin{figure}[t] 
	\centering  
	\includegraphics[width=1.0\linewidth]{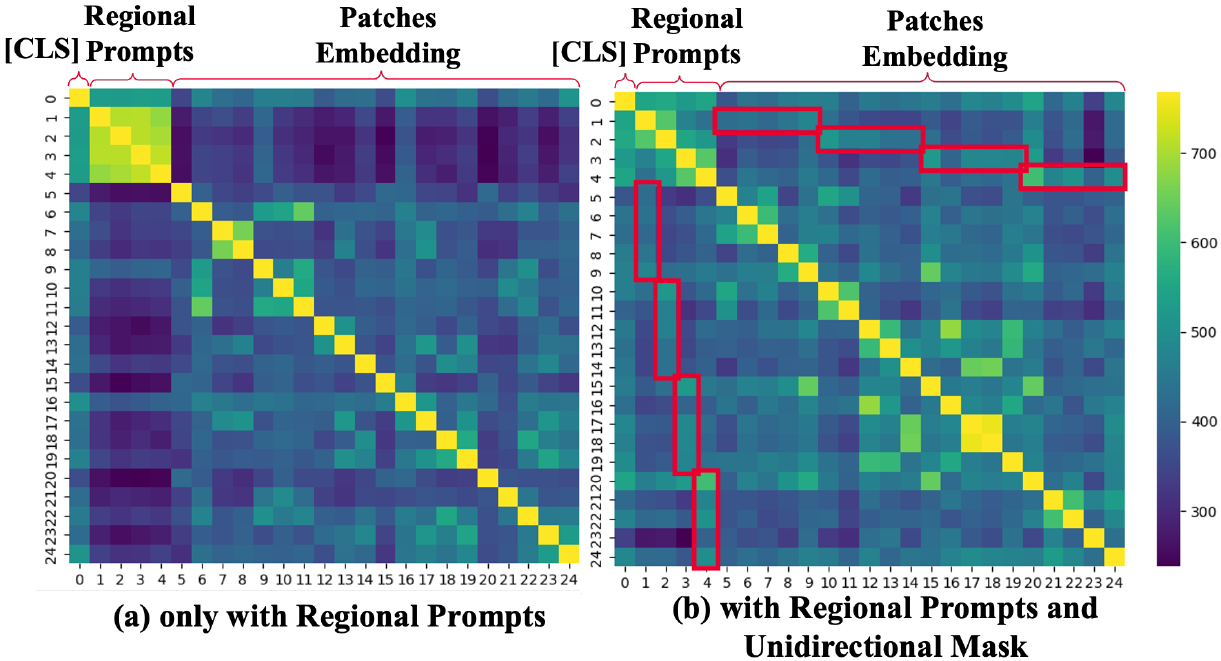}
	\caption{Visualization of the Effects of Unidirectional Masking and Region Prompts.}
	\label{fig:maps}
\end{figure}

In \cref{fig:maps}, the regional prompts obtain stronger responses in the corresponding local patches. The red boxes visualize how regional prompts incorporate local features, highlighting the role of Unidirectional Mask. Moreover, the heatmap (b) exhibits higher global responses compared to heatmap (a).

\begin{figure*}[t]
	\centering  
        \begin{subfigure}{1.\linewidth}
            \centering
            \includegraphics[width=0.95\linewidth]{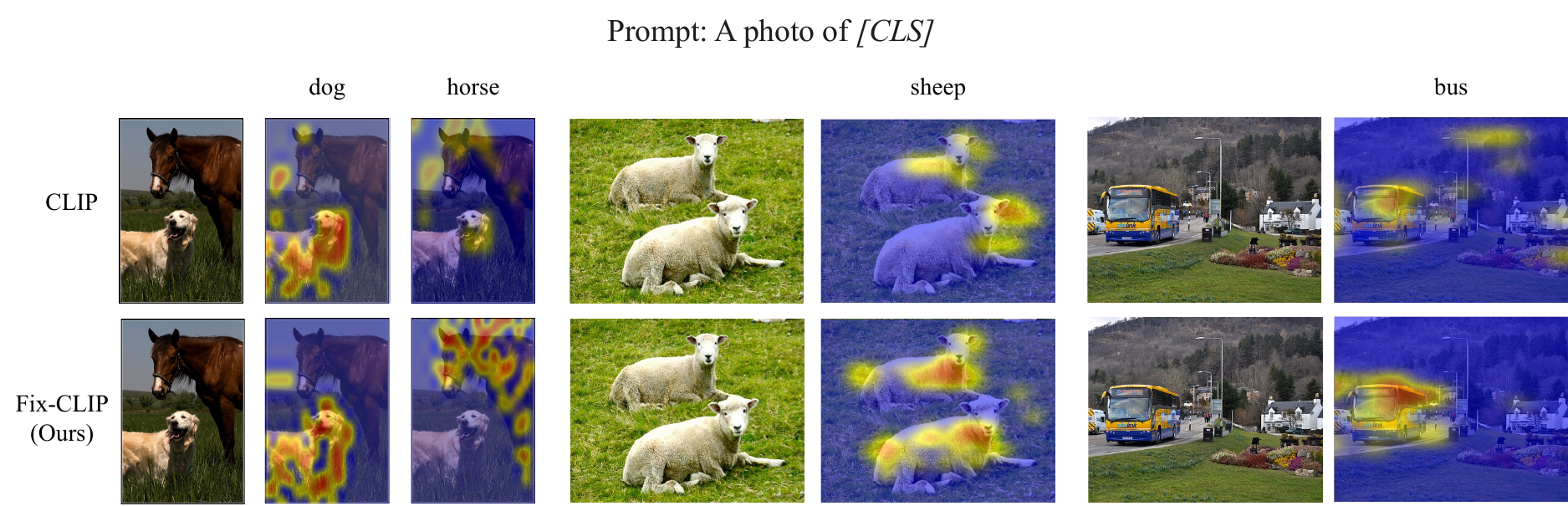}
            \caption{Similarity heatmap compared with CLIP~\cite{radford2021learning}.}
            \label{fig:append_heatmap_a}
        \end{subfigure}
        
        \vspace{1.0em}
        
        \begin{subfigure}{1.\linewidth}
            \centering
            \includegraphics[width=0.95\linewidth]{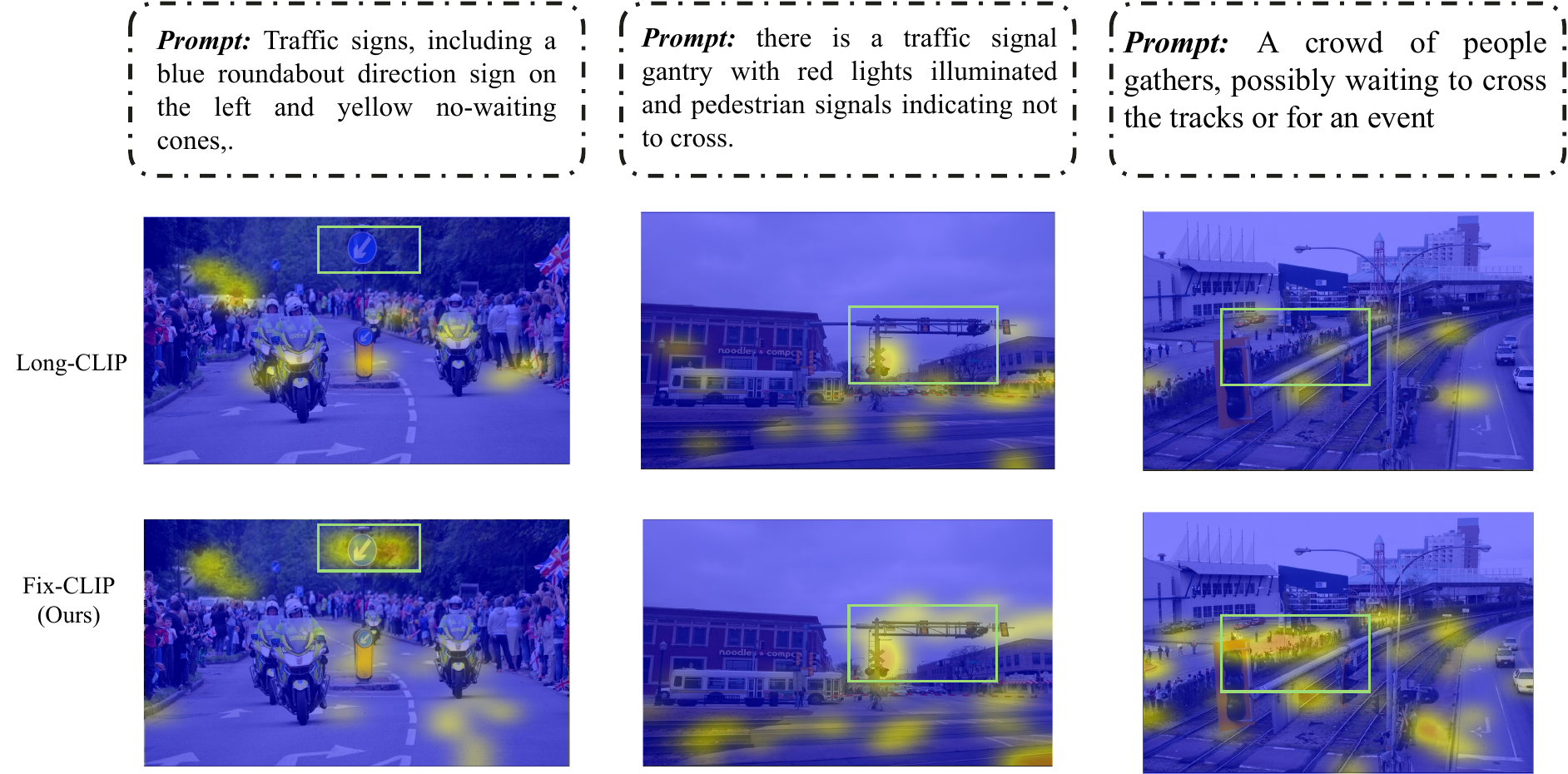}
            \caption{Similarity heatmap compared with Long-CLIP~\cite{zhang2024long}.}
            \label{fig:append_heatmap_b}
        \end{subfigure}
	\caption{Similarity Heatmap between text and image features in different models. (a) presents a comparative analysis between our model and CLIP~\cite{radford2021learning} in short-text scenarios, while (b) illustrates the performance comparison between our model and Long-CLIP~\cite{zhang2024long} in long-text contexts. The text segments highlighted in red represent semantic information successfully comprehended by our model but not accurately captured by Long-CLIP~\cite{zhang2024long}.}
	\label{fig:append_heatmap}
\end{figure*}

\section{Visualization of the Similarity Heatmap}
\label{sec: heatmap}
We visualize the heatmap of similarity between image features and text features, and compare our results with those of CLIP~\cite{radford2021learning} and Long-CLIP~\cite{zhang2024long}, as shown in \cref{fig:append_heatmap}. To evaluate the performance on short texts, the prompt is set as "a photo of \textit{[CLS]}". \ours demonstrates superior performance over CLIP~\cite{radford2021learning}, accurately identifying instances in the image, as illustrated in \cref{fig:append_heatmap_a}. For long-text understanding, the prompt consists of a major sentence split from the original long-text captions, enabling a direct comparison with Long-CLIP~\cite{zhang2024long}. The corresponding performance is depicted in \cref{fig:append_heatmap_b}.

\section{Analysis of Text-to-Image Generation Examples}
\label{sec: more generation examples}
\begin{figure*}[ht] 
	\centering  
	\includegraphics[width=0.8\linewidth]{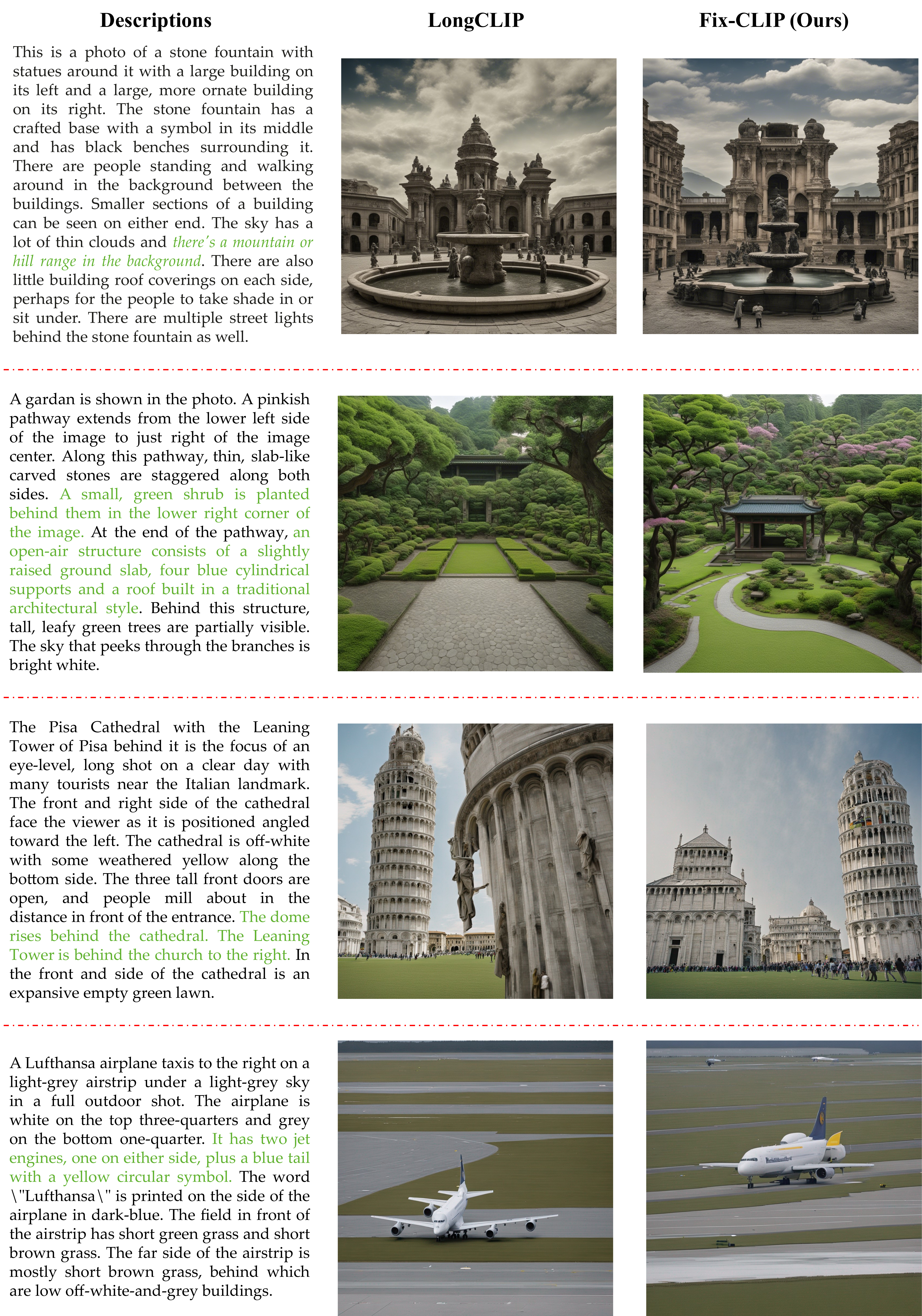}
	\caption{More Text-to-Image Generation examples. Images generated by \ours are more accurate in detail information such as color, direction, position, quantity, material, light, and shooting angle. The text highlighted in green represents fine-grained details that Long-CLIP~\cite{zhang2024long} fails to capture, whereas our proposed model \ours successfully generates these contextual elements with high fidelity.}
	\label{fig:dataset1}
\end{figure*}

In this section, we showcase more text-to-image generation examples in long captions to demonstrate the enhancement in understanding long texts. We replace the original text encoder in the stable-diffusion model with that in Long-CLIP~\cite{zhang2024long} or ours. Then, the reconstructed model would be fed with the long captions in the~\cite{garg2024imageinwords} dataset. Due to the divergence between the original text encoder and our text encoder, the model is restrained to generate coarse images. Therefore, an image-to-image refiner model is utilized subsequently to transfer the coarse images to fine images. The final performance is illustrated in \cref{fig:dataset1}. The result of Long-CLIP~\cite{zhang2024long} has confusion in some details, \ie the background, the direction, and the position relation. Even hallucinations would occur, such as the airplane equipping four jet engines in the $4$-th case. For the comparison, our model correctly describes the detailed information and performs better.

\end{document}